\documentclass[10pt,twocolumn,letterpaper]{article}
\PassOptionsToPackage{dvipsnames,svgnames,x11names}{xcolor}
\PassOptionsToPackage{inline}{enumitem}
\usepackage[numbers,sort&compress]{natbib}

\usepackage{thunlp_cvpr}

\usepackage[T1]{fontenc}

\usepackage[utf8]{inputenc}

\usepackage{microtype}

\usepackage{inconsolata}

\usepackage{graphicx}

\usepackage{subcaption}
\usepackage{booktabs} 

\usepackage{hyperref}

\usepackage{mathtools}

\usepackage{amsthm}

\usepackage{url}
\usepackage{amsfonts} 
\usepackage{nicefrac} 
\usepackage[dvipsnames, svgnames, x11names]{xcolor}

\usepackage{multirow}
\usepackage{amsmath}
\usepackage{changepage}

\makeatletter

\let\algorithmic@indent\relax
\makeatother

\usepackage{algpseudocode}
\usepackage{algorithm}

\usepackage[inline]{enumitem}

\usepackage{wrapfig}

\usepackage{colortbl}
\usepackage{tcolorbox}
\usepackage{enumitem}
\usepackage{makecell}
\usepackage{tabularx}        
\tcbuselibrary{breakable,skins}

\tcbuselibrary{breakable}

\usepackage{listings}
\definecolor{codegreen}{rgb}{0,0.6,0}
\definecolor{codegray}{rgb}{0.5,0.5,0.5}
\definecolor{codepurple}{rgb}{0.58,0,0.82}
\definecolor{backcolour}{rgb}{0.95,0.95,0.92}
\definecolor{panelgray}{RGB}{242,242,242} 

\newtcolorbox{casestudy}[2][]{%
  enhanced,
  breakable,
  colback=black!1,          
  colframe=black!60,        
  boxrule=0.4pt,
  arc=2pt,                  
  left=10pt,right=10pt,top=10pt,bottom=10pt,
  title={#2},
  fonttitle=\bfseries\scshape, 
  colbacktitle=white,
  coltitle=black,
  boxed title style={boxrule=0.4pt, colframe=black!60, sharp corners},
  before skip=6pt, after skip=6pt, 
  #1
}

\lstdefinestyle{mystyle}{
    commentstyle= \color{red!50!green!50!blue!50},  
    keywordstyle= \color{blue!70},  
    numberstyle=\tiny\color{codegray},  
    stringstyle=\color{codepurple},
    basicstyle=\ttfamily\footnotesize,
    breakatwhitespace=false,
    breaklines=true,  
    captionpos=b,
    keepspaces=true,
    showspaces=false,
    showstringspaces=false,  
    showtabs=false,
    tabsize=2,
    xleftmargin=4pt, xrightmargin=4pt,
    frame=single  
}

\usepackage[capitalize,noabbrev]{cleveref}

\theoremstyle{plain}

\theoremstyle{definition}

\theoremstyle{remark}

\usepackage[disable,textsize=tiny]{todonotes}

\makeatletter
\newcommand{\ie}{\emph{i.e.}\@ifnextchar.{\!\@gobble}{}}
\newcommand{\eg}{\emph{e.g.}\@ifnextchar.{\!\@gobble}{}}
\newcommand{\etc}{etc\@ifnextchar.{}{.\@}}
\makeatother

\title{Agent-Environment Alignment via\\Automated Interface Generation}

\author{
 \textbf{Kaiming Liu\textsuperscript{1,2}},
 \textbf{Xuanyu Lei\textsuperscript{2,3}},
 \textbf{Ziyue Wang\textsuperscript{2,3}},
 \textbf{Peng Li\textsuperscript{2,$\dagger$}},
 \textbf{Yang Liu\textsuperscript{1,2,3,$\dagger$}} \\
 \thunlpauthoraffildivider
}
\thunlpaffiliation{
 \textsuperscript{1}College of AI, Tsinghua University, Beijing, China\\
 \textsuperscript{2}Institute for AI Industry Research (AIR), Tsinghua University, Beijing, China\\
 \textsuperscript{3}Dept. of Comp. Sci. \& Tech., Institute for AI, Tsinghua University, Beijing, China
}

\newcommand{\alignabstractnote}{Code and data are available at \url{https://github.com/THUNLP-MT/ALIGN}.}
\usepackage{xurl}
\usepackage{adjustbox}
\usepackage{needspace}
\usepackage{flushend}
\thunlplogooff
\setthunlparticlemeta{RESEARCH ARTICLE}
\date{}

\makeatletter
\def\@maketitle{%
  \vspace*{-0.7em}
  \noindent
  \begin{tikzpicture}[remember picture,overlay]
    \node[anchor=north east,inner sep=0pt]
      at ([xshift=-0.75in,yshift=-0.67in]current page.north east)
      {\includegraphics[width=0.23\textwidth]{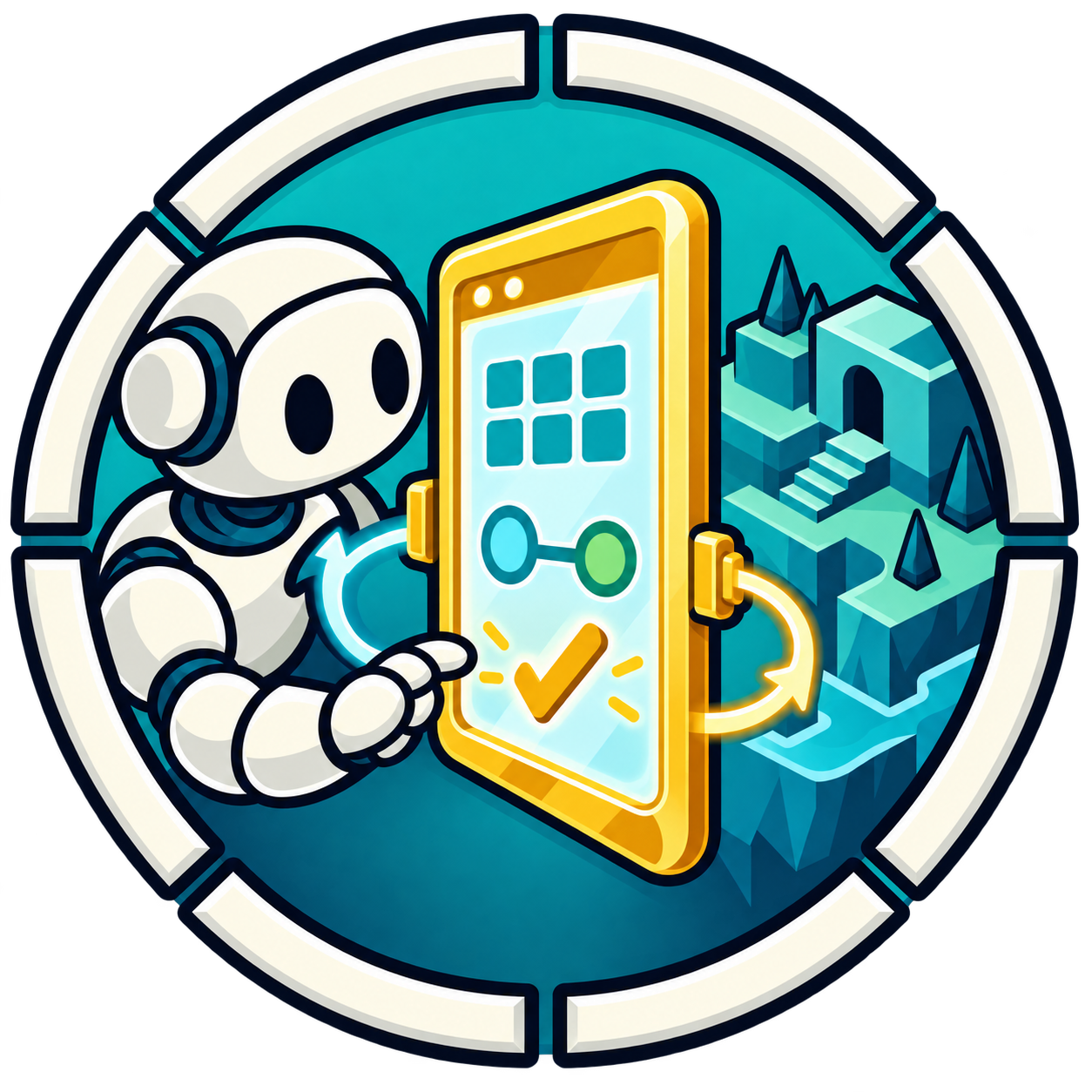}};
  \end{tikzpicture}%
  \begin{minipage}[t]{0.74\textwidth}
    \vspace{0pt}
    {\setlength{\fboxsep}{5pt}\colorbox{THUDarkTint}{\textbf{\small\color{THUAccent}\thunlpArticleType}}}\par
    \vspace{0.55em}
    {\fontsize{20}{24}\selectfont\bfseries\color{THUDark}\raggedright\@title\par}
  \end{minipage}
  \par\vspace{0.55em}
  {\normalsize\@author\par}
  \ifx\thunlp@affiliation\@empty\else
    \par\vspace{0.24em}
    {\normalsize\parbox[t]{\textwidth}{\thunlp@affiliation}\par}
  \fi
  \par\vspace{0.45em}
  \@thanks
  \thispagestyle{fancy}
}
\makeatother

\renewcommand{\makefrontmatter}[1]{%
  \twocolumn[
  \maketitle
  \begin{abstract}
    #1\unskip\textsuperscript{a}
    \par\vspace{0.35em}
    \noindent\rule{0.38\linewidth}{0.35pt}\par
    \vspace{0.1em}
    {\fontsize{8}{10}\selectfont\urlstyle{rm}
      \noindent\textsuperscript{a}\alignabstractnote\par}
  \end{abstract}
  \vspace{0.25em}
  ]
}

\hypersetup{
  pdftitle={Agent-Environment Alignment via Automated Interface Generation},
  pdfauthor={Kaiming Liu, Xuanyu Lei, Ziyue Wang, Peng Li, Yang Liu},
  pdfsubject={ALIGN in the THUNLP conference-style template},
  pdfkeywords={ALIGN, agent-environment alignment, automated interface generation},
  bookmarksnumbered=true
}
\tcbset{alignprompt/.style={
  enhanced,breakable,
  colback=THULightTint,colframe=THUAccent,
  colbacktitle=THUAccent,coltitle=white,
  fonttitle=\bfseries\normalsize,
  fontupper=\small,
  boxrule=0.6pt,arc=2pt,
  left=9pt,right=9pt,top=8pt,bottom=8pt,
  before skip=12pt,after skip=12pt,
  before upper={\raggedright\setlength{\parindent}{0pt}\setlength{\parskip}{2pt}},
  pad at break*=2mm
}}

\begin{document}
\makefrontmatter{
Large language model agents have shown impressive capabilities in interactive decision-making tasks.
These agents interact with environment through intermediate interfaces, such as predefined action spaces and interaction rules, which mediate the perception and action.
However, mismatches often happen between the internal expectations of the agent regarding the influence of its issued actions and the actual state transitions in the environment, a phenomenon referred to as \textbf{agent-environment misalignment}.
While prior work has invested substantially in improving agent strategies and environment design, the critical role of the interface still remains underexplored.
In this work, we empirically demonstrate that agent-environment misalignment poses a significant bottleneck to agent performance.
To mitigate this issue, we propose \textbf{ALIGN}, an \underline{A}uto-A\underline{l}igned \underline{I}nterface \underline{G}e\underline{n}eration framework that alleviates the misalignment by enriching the interface.
Implemented as a lightweight wrapper, ALIGN-generated interface achieves the alignment without modifying either the agent logic or the environment code.
Experiments across multiple domains including embodied tasks, web navigation and tool-use, achieve consistent performance improvements, with a 45.67\% success rate improvement in ALFWorld.
}

\begingroup
\renewcommand{\thefootnote}{}
\footnotetext{\urlstyle{rm}$^\dagger$ Corresponding Authors: \\
\href{mailto:lipeng@air.tsinghua.edu.cn}{\nolinkurl{lipeng@air.tsinghua.edu.cn}},
\href{mailto:liuyang2011@tsinghua.edu.cn}{\nolinkurl{liuyang2011@tsinghua.edu.cn}}.}
\endgroup
\setcounter{footnote}{0}

\section{Introduction}
\label{sec:introduction}

Large Language Model (LLM) agents have demonstrated promising performance in interactive tasks such as embodied tasks~\citep{DBLP:journals/tmlr/WangX0MXZFA24}, web navigation tasks~\citep{DBLP:journals/corr/abs-2411-02337}, and tool-use tasks~\citep{DBLP:conf/nips/SchickDDRLHZCS23}.
In these tasks, agents typically interact with the environment through manually designed \textbf{interfaces} such as predefined action spaces and interaction rules.
While substantial efforts have been devoted to improving agents and environments, comparatively little attention has been paid to the interface between them, leading to a problem we term \textbf{agent-environment misalignment}, which impacts the agent performance.

\begin{figure*}[t]
        \centering
        \includegraphics[width=\linewidth]{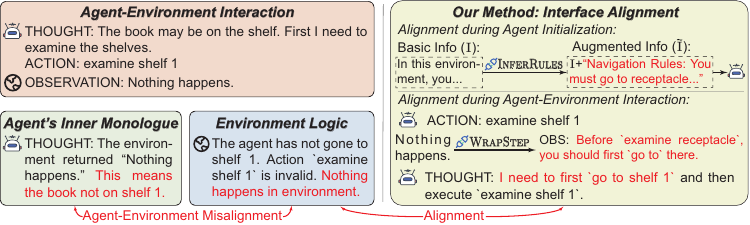}
    \caption{\textbf{Illustration of agent-environment misalignment and our proposed solution.} \textbf{Left:}~The agent and the environment have a misalignment in their interpretation of the same observation, where the agent's understanding of the observation differs from the environment's underlying logic. \textbf{Right:}~ALIGN bridges this gap via an automatically generated interface comprising two modules, \textbf{\textsc{InferRules}} and \textbf{\textsc{WrapStep}}: 
    (1)~During initialization, \textsc{InferRules} augments the basic information with explicit environment constraints.
    (2)~During interaction, \textsc{WrapStep} translates the raw observation into an informative observation.}
    \label{fig:first}
\end{figure*}

The agent-environment misalignment refers to the discrepancy between the interpretation of the agent to the observation following an action and the underlying logic of the environment.
As illustrated in Figure~\ref{fig:first} (left), in ALFWorld~\citep{DBLP:conf/iclr/ShridharYCBTH21}, an invalid action returns ``Nothing happens'', which agents misinterpret as an empty state rather than a constraint violation.
This misalignment significantly hinders task success and imposes unnecessary cognitive overhead on agents.
From the perspective of the agent, poorly designed interfaces impose unnecessary cognitive overhead.
Furthermore, from an evaluation perspective, inadequate interfaces can obscure an accurate assessment of the true reasoning capabilities of agents.
Therefore, we argue that the problem of agent-environment misalignment warrants greater attention.

However, addressing the agent-environment misalignment is challenging.
On one hand, current benchmarks primarily focus on advance agent intelligence by constructing increasingly complex and challenging environments~\citep{DBLP:conf/nips/XieZCLZCHCSLLXZ24,DBLP:conf/iclr/ZhouX0ZLSCOBF0N24}, often overlooking the importance of improving interface design.
This oversight extends across multiple domains of interactive tasks, such as, ALFWorld, OSWorld~\citep{DBLP:conf/nips/XieZCLZCHCSLLXZ24}, and M$^3$ToolEval~\citep{DBLP:conf/icml/WangCY0L0J24}. They all exhibit similar deficiencies: failing to provide agent-parseable observations for environmental constraints violation in embodied tasks, positional inaccuracies in operating system tasks or parameter format errors in multi-turn tool-use tasks, respectively.
On the other hand, although some recent work~\citep{DBLP:conf/nips/YangJWLYNP24,DBLP:conf/icml/ZhengGK0024} has begun to consider interface design, these efforts often rely on manual, environment-specific tailoring, which introduces two critical issues: (1) they are highly labor-intensive and (2) whether human-designed interfaces are optimal for agents remains an open question.
Furthermore, researchers often manually re-engineer interfaces for specific methods~\citep{DBLP:conf/nips/MaZZYYJLKH24, DBLP:conf/nips/0001LYYL024}, compromising comparability and making it unclear whether improvements stem from novel architectures or customized interfaces.

Distinct from the aforementioned works, we propose to \textbf{automatically generate interfaces for bridging the agent-environment misalignment}.
In this work, we introduce \textbf{ALIGN} (\underline{A}uto-A\underline{l}igned \underline{I}nterface \underline{G}e\underline{n}eration), a framework that automatically generate aligned interfaces for environments.
The generated interface consists of two modules: \textsc{InferRules} and \textsc{WrapStep}.
The former discovers and provides the agent with static information about environmental rules or internal constraints, facilitating \textit{static alignment}, while the latter enhances the interaction by offering more detailed observations for agent-issuing actions, enabling \textit{dynamic alignment}, as shown in Figure~\ref{fig:first} (right).
The generated interface acts as a lightweight wrapper, providing richer context and explicit constraint hints, enabling different LLM agents to align with the environment directly.

To evaluate the effectiveness of ALIGN, we conduct experiments on four representative benchmarks across three domains: embodied tasks, web navigation, and tool-use tasks.
Our results demonstrate consistent performance improvements across all four benchmarks when using the ALIGN-generated interface, with notably gains of 45.67\% in average success rate on ALFWorld.
Moreover, the performance of GPT-4.1-based agents on ALFWorld are improved from 73.88\% to 93.28\% with ALIGN, highlighting the efficiency of our approach in mitigating the agent-environment misalignment to unleash the agent’s true capabilities.

Our contributions are summarized as follows:
\begin{itemize}[leftmargin=1.5em,itemsep=0pt,parsep=0.2em,topsep=0.1em,partopsep=0.0em]
    \item We identify and characterize the \textbf{agent-environment misalignment}, empirically showing its prevalence across diverse domains and its role as a key bottleneck to performance.
    \item We introduce \textbf{ALIGN}, the first framework that automatically generates aligned interfaces to alleviate agent-environment misalignment, without modifying agent logic or environment code.
    \item We demonstrate the effectiveness and generalizability of \textbf{ALIGN} across three domains, with up to a 45.67\% success rate gain on ALFWorld.
\end{itemize}

\section{Related Work}
\label{sec:related}
\textbf{Agent-environment interface}~
The agent-environment interface defines how agents interact with the environment.
In reinforcement learning, unified interaction interfaces~\citep{DBLP:conf/iclr/BonnetLBSADCMTK24,DBLP:journals/corr/BrockmanCPSSTZ16, DBLP:journals/corr/abs-1712-05474, DBLP:journals/corr/abs-2407-17032} standardize the evaluation of different algorithms.
As LLMs become increasingly capable of human-like actions~\citep{DBLP:conf/ijcai/GuoCWCPCW024, DBLP:conf/iclr/0036YZXLL0DMYZ024, DBLP:conf/nips/MaZZYYJLKH24}, interface design has been shown to largely influence LLM-based agent performance~\citep{DBLP:conf/nips/XieZCLZCHCSLLXZ24, DBLP:journals/corr/abs-2405-14573}. 
SWE-agent~\citep{DBLP:conf/nips/YangJWLYNP24} proposes agent-computer interfaces for coding agents, and recent efforts aim to improve generalization~\citep{DBLP:journals/corr/abs-2410-08164,DBLP:journals/corr/abs-2501-12326,DBLP:conf/ijcai/NiuL0FHLKCW24} or augment interfaces with auxiliary tools~\citep{bula2025seaview,DBLP:journals/corr/abs-2410-05243,DBLP:conf/coling/LeiYCLL25,DBLP:journals/corr/abs-2408-00203,DBLP:journals/corr/abs-2310-11441}.
However, these interfaces are mostly hand-crafted for specific environments or frameworks, limiting their generalization and scalability.
We instead propose automated interface generation for effective, generalizable alignment.

\textbf{Methods aligning agents with environments}~
LLM agents have shown strong potential for real-world interaction and task completion~\cite{DBLP:conf/iclr/YaoZYDSN023, DBLP:conf/nips/ShinnCGNY23, DBLP:conf/iclr/0036YZXLL0DMYZ024}, with existing work falling into training-based and training-free categories.
Training-based methods fine-tune LLMs on expert trajectories~\cite{DBLP:conf/acl/ZengLLWLD024, DBLP:journals/corr/abs-2310-05915, DBLP:journals/corr/abs-2503-02197, DBLP:journals/corr/abs-2501-01702, DBLP:conf/acl/ChenLWZLLCZ24} or apply reinforcement learning for environment-aligned planning~\cite{bai2025digiq, DBLP:journals/corr/abs-2403-14589, DBLP:journals/corr/abs-2411-02337, DBLP:conf/nips/FengHHLZZL24, DBLP:conf/icml/ZhouZPLK24, ragen}, but suffer from high computational costs and poor generalization to unseen environments.
Training-free methods build multi-agent frameworks for task decomposition and experience accumulation~\citep{DBLP:conf/nips/0001LYYL024,he2024ideaenhancingrulelearning,DBLP:journals/corr/abs-2412-19723,DBLP:conf/emnlp/YangLL23,DBLP:journals/corr/abs-2410-07484}, yet their static pipelines and prompt-injected experience often fail to capture environment dynamics, yielding insufficient agent-environment alignment.

\textbf{Rule learning and prompt optimization}~
Closest to our work are methods that learn \emph{agent-side} knowledge: WALL-E~\citep{DBLP:journals/corr/abs-2410-07484} induces rules that correct the agent's world model, AutoManual~\citep{DBLP:conf/nips/0001LYYL024} builds a manual that conditions its planner, and automatic prompt optimization (APO) methods~\citep{zhou2023ape,pryzant2023protegi,yang2024opro,wang2024promptagent} refine the agent's prompt with an LLM-driven search loop.
ALIGN instead optimizes the environment-side functions $f_{\text{info}}$ and $f_{\text{obs}}$, produces executable interface code rather than natural-language knowledge, and accepts a change only after controlled experiments in the environment.
Since these methods leave the feedback channel untouched, they consume exactly the raw observations that misalignment corrupts and are thus complementary to ALIGN (Section~\ref{sec:further_analyses}); a detailed comparison is given in Appendix~\ref{app:positioning}.

\section{Method}
\label{sec:method}

\subsection{Problem Formulation}
\label{sec:3.1}

\textbf{Environment and Agent.}~
In interactive decision tasks, we define the environment $\mathcal{E}$ as a tuple $(\mathcal{S}, \mathcal{A}, T, F, \mathcal{I})$, where $\mathcal{S}$ denotes the set of all possible states of the environment; $\mathcal{A}$ denotes the set of actions the agent can invoke; $T: \mathcal{S} \times \mathcal{A} \rightarrow \mathcal{S}$ defines how the environment state evolves in response to agent actions; $F: \mathcal{S} \times \mathcal{A} \rightarrow \mathcal{O}$ provides textual feedback that reflects the consequences of the action in the current state, where $\mathcal{O}$ is all possible observations; $\mathcal{I}$ encodes the \textit{environment foundational information description}, a fixed, declarative representation of the environment's basic introduction, which is exposed to the agent at initialization.
An agent $\pi$ interacts with environment $\mathcal{E}$ by receiving task description and observations, then producing actions $a_t \in \mathcal{A}$. The interaction generates trajectory $\tau = [(s_0, a_0, o_0), \ldots, (s_T, a_T, o_T)]$, culminating in task completion feedback.

\textbf{Formal Definition of Interface.}~
Existing works typically assume the agent interacts directly with $\mathcal{E}$. 
However, we argue that this interaction is mediated by an \textbf{Interface}, denoted as $\Phi$, which acts as a translation layer between the agent's cognitive space and the environment's execution space. Formally, we define the interface as a tuple of mapping functions: \(\Phi = \langle f_{\text{info}}, f_{\text{act}}, f_{\text{obs}} \rangle\)where each component serves a distinct role:

\begin{itemize}[leftmargin=1.5em,itemsep=0pt,parsep=0.2em,topsep=0.1em,partopsep=0.0em]
    \item \textbf{Information Augmenter} $f_{\text{info}}: \mathcal{I} \rightarrow \tilde{\mathcal{I}}$ exposes implicit environment logic (e.g., admissible action sequences) into an explicit descriptive context $\tilde{\mathcal{I}}$ provided at agent initialization.
    \item \textbf{Action Transducer} $f_{\text{act}}: \mathcal{A}_{\text{agent}} \rightarrow \mathcal{A}_{\text{env}} \cup \{\perp\}$ maps the agent's output to an executable environment command. If the output cannot be parsed, it returns an invalid symbol $\perp$.
    \item \textbf{Observation Transducer} $f_{\text{obs}}: \mathcal{S} \times \mathcal{A}_{\text{env}} \times \mathcal{O}_{\text{raw}} \rightarrow \mathcal{O}_{\text{agent}}$ transforms the raw feedback $\mathcal{O}_{\text{raw}}$ (from $F$) into an informative observation $\mathcal{O}_{\text{agent}}$ that better conveys the actual state transitions and their causes.
\end{itemize}

At each timestep $t$, the agent receives $\tilde{o_t} \in \mathcal{O}_{\text{agent}}$ (processed by $f_{\text{obs}}$) and generates $a_t \in \mathcal{A}_{\text{agent}}$, which is then executed as $a^{\text{env}}_t = f_{\text{act}}(a_t) \in \mathcal{A}_{\text{env}}$.

\begin{figure*}[t]
    \centering
    \includegraphics[width=\linewidth]{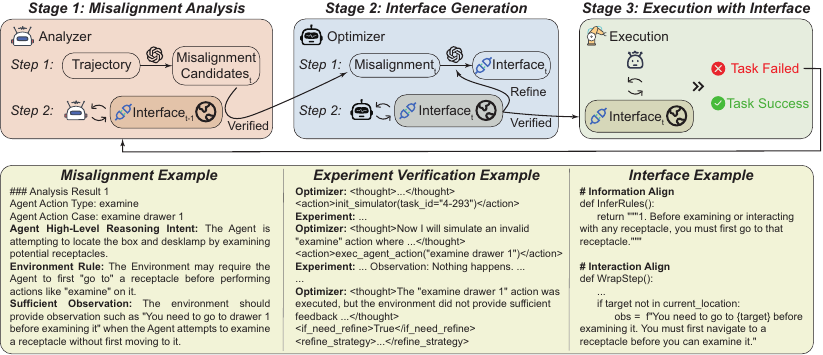}
    \caption{\textbf{ALIGN framework.} In each iteration, ALIGN progresses though three stages. \textbf{Stage~1}: the Analyzer identifies potential agent-environment misalignments and validates them through experiments; \textbf{Stage~2}: the Optimizer generates a new interface based on the previous interface and identified misalignments, followed by verification and refinement; \textbf{Stage~3}: the agent interacts with the updated interface-wrapped environment, with trajectories of failed tasks fed back to the Analyzer for analysis in the next iteration. At the bottom of the figure, examples for misalignment, verification of interface by Optimizer, and the ALIGN-generated interface are provided.}
    \label{fig:system}
\end{figure*}

\textbf{Agent-Environment Misalignment.}
We analyze the misalignment problem through the lens of the interface $\Phi$. 
Ideally, $\Phi$ should be \textit{lossless}, maximizing the mutual information between the agent's belief state and the ground-truth environment state. However, manually designed interfaces often exhibit \textbf{Semantic Gaps} through two mechanisms: (1) \textit{state aliasing}, where a lossy $f_{\text{obs}}$ maps distinct error states (e.g., wrong location vs. missing object) to the same generic observation (e.g., ``Nothing happens.''), preventing the agent from diagnosing failures; and (2) \textit{under-specified constraints}, where critical preconditions of $T$ (e.g., ``\texttt{open}'' requires ``\texttt{go to}'' first) are not encoded in $\tilde{\mathcal{I}}$ by $f_{\text{info}}$, leaving the agent under a hallucinated world model.

Therefore, we define \textbf{Agent-Environment Misalignment} as the discrepancy between the agent's expected state transition $s^{\text{expected}}_{t+1}$ (derived from its internal world model based on $\tilde{\mathcal{I}}$ and prior observations $[\tilde{o_0}, \tilde{o_1}, \ldots, \tilde{o_{t+1}}]$) and the actual transition $s^{\text{actual}}_{t+1} = T(s_t, a^{\text{env}}_t)$, caused by insufficient expressiveness of the interface $\Phi$.

\textbf{Scope of This Work.}
We observe that misalignment in existing benchmarks primarily stems from information loss in $f_{\text{info}}$ and $f_{\text{obs}}$ rather than action space incompatibility. ALIGN therefore treats $f_{\text{act}}$ as the identity mapping and focuses on automatically optimizing the other two components. An audit of all 87 \texttt{env.step} call sites in the 20 generated interfaces supports this choice empirically: none remaps an agent action to a semantically different one (Appendix~\ref{app:fact_audit}).

\subsection{ALIGN Overview}

We introduce \textbf{ALIGN}~(\underline{A}uto-A\underline{l}igned \underline{I}nterface \underline{G}e\underline{n}eration), a framework that automatically generates an optimized interface $\Phi^*$ to bridge the semantic gaps identified in Section~\ref{sec:3.1}. As illustrated in Figure~\ref{fig:interface}, ALIGN instantiates $f_{\text{info}}$ and $f_{\text{obs}}$ as two learnable modules implemented as a lightweight Python wrapper:

\begin{figure}[htbp]
    \begin{adjustwidth}{0.6in}{0.6in}  
        \centering
        \includegraphics[width=\linewidth]{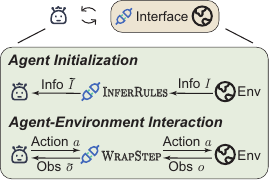}
    \end{adjustwidth}
    \caption{Overview of the ALIGN-generated interface. \textsc{InferRules} enriches static information ($\mathcal{I} \rightarrow \tilde{\mathcal{I}}$) at agent initialization, while \textsc{WrapStep} augments step-wise observations ($F \rightarrow \tilde{F}$) during interaction.}
    \label{fig:interface}
\end{figure}

\textbf{Interface Instantiation.}~
As illustrated in Figure~\ref{fig:interface}, ALIGN instantiates the theoretical interface components through two learnable modules implemented as a lightweight Python wrapper, without modifying the environment code or agent logic.

\textbf{\textsc{InferRules}} (implements $f_{\text{info}}$) extracts interaction rules and constraints from $(\text{task}, o_0, \mathcal{I})$ to produce $\tilde{\mathcal{I}}$, exposing precondition dependencies and action ordering requirements at initialization.

\textbf{\textsc{WrapStep}} (implements $f_{\text{obs}}$) intercepts $F$ and augments each raw observation with diagnostic information to resolve state aliasing in interaction.

Together, these modules wrap $\mathcal{E}$ into an \textit{augmented environment} $\tilde{\mathcal{E}} = (\mathcal{S}, \mathcal{A}, T, \tilde{F}, \mathcal{I}\cup\tilde{\mathcal{I}})$ with $\tilde{F}(s_t, a_t) := \textsc{WrapStep}(F, s_t, a_t)$, leaving the agent logic, the environment code, and the transition dynamics $T$ unchanged. We denote the interface as $\Phi := \{\textsc{InferRules}, \textsc{WrapStep}\}$ in the remainder of this paper.

As shown in Figure~\ref{fig:system}, the ALIGN integrates two cooperative modules, \textbf{Analyzer} and \textbf{Optimizer}, to generate aligned interfaces.
The framework operates through iterative optimization, with each iteration comprising three stages: in Stage~1, the Analyzer identifies agent-environment misalignments by analyzing past interaction trajectories; in Stage~2, the Optimizer generates, validates and refines a new interface based on the detected misalignments; and in Stage~3, the agent interacts with the environment wrapped with the newly generated interface, and the failed task trajectories are fed back to Analyzer for analysis in the next iteration.

\subsection{ALIGN Framework\label{sec:ALIGN_framework}}

To automate the generation of interfaces that bridge the agent-environment misalignments, ALIGN need to solve two key challenges: (1)~how to analyze and identify existing agent-environment misalignments, and (2)~how to generate an interface that addresses these misalignments. The overall algorithm process of ALIGN is outlined in Algorithm~\ref{alg:align} in Appendix~\ref{app:formalization}.

\textbf{Misalignment Analysis.}~
We represent each agent-environment misalignment using structured text, as shown in the bottom left of Figure~\ref{fig:system}.
The ``Agent High-Level Reasoning Intent'' and ``Environment Rule'' respectively depict the agent's expectations of the action and the environment's observation rules, together representing a misalignment.
The ``Sufficient Observation'' represents the observation the environment should provide to resolve the misalignment.
To analyze and identify these misalignments, we designed the Analyzer module based on LLMs.
In each iteration, the Analyzer takes the failed interaction trajectory $\tau^{(i-1)}$ in the previous iteration, the set of currently identified misalignments $\mathcal{M}$, and the interface $\Phi^{(i-1)}$ from the previous round as input, and generates a new set of misalignments $\mathcal{M}^{(i)}$.
The Analyzer needs no human labels or gold trajectories; its supervision is the agent's failed trajectories and the environment's outcome signal (Appendix~\ref{app:prompt}).
Detailed prompts for this process are provided in Appendix~\ref{app:prompt}.

\textbf{Interface Generation.}~
Once the new set of misalignments $\mathcal{M}^{(i)}$ is identified, we employ the Optimizer module to generate a new interface.
We represent the two modules of the interface, \textsc{InferRules} and \textsc{WrapStep}, as Python functions, as shown in the bottom right of Figure~\ref{fig:system}, to leverage the powerful code generation capabilities of LLMs.
In each iteration, the Optimizer takes the newly identified misalignments $\mathcal{M}^{(i)}$ and the previous interface $\Phi^{(i-1)}$ as input, and generates a new interface $\Phi^{(i)}$.
The detailed prompts for this process are provided in Appendix~\ref{app:prompt}.

\textbf{Experimental Verification.}~
Given the hallucination~\citep{DBLP:conf/ijcnlp/BangCLDSWLJYCDXF23,DBLP:journals/corr/abs-2401-11817} issues of LLMs, we incorporate an experimental verification procedure.
Specifically, after the Analyzer generates $\mathcal{M}^{(i)}$, it will interact with the environment wrapped by the previous interface $\Phi^{(i-1)}$ to validate whether the identified misalignments do indeed exist and can be resolved by the proposed ``Sufficient Observation''.
And after the Optimizer generates the new interface $\Phi^{(i)}$, it will interact with the environment wrapped by this new interface to ensure that the generated interface can resolve the identified misalignments.
If the Optimizer finds that the proposed interface is insufficient to address the discovered misalignments, it will provide a refinement strategy and regenerate the interface.
This iterative process continues until the interface passes the validation, ensuring that the misalignments identified are appropriately addressed.
An example of this process is provided in the bottom center of Figure~\ref{fig:system}.
To facilitate this interaction with the interface-wrapped environment, we designed a set of encapsulated tools for both the Analyzer and Optimizer to use, as described in Appendix~\ref{app:apis}.

After each iteration, the agent interacts with the environment wrapped by the new generated interface $\Phi^{(i)}$, and trajectories of the failed tasks are returned to Analyzer for further analysis.
The algorithm continues iteratively until one of the following holds: (1)~the pre-defined maximum number of iterations is reached; (2)~no failed trajectories are produced; (3)~no new misalignments are identified.

\begin{table*}[t]
    \centering
    \begingroup
        \fontsize{8.5}{10}\selectfont
        \setlength{\tabcolsep}{4pt}
        \renewcommand{\arraystretch}{1}
        \begin{adjustbox}{max width=\linewidth}
        \begin{tabular}{@{}llcccc@{}}
            \toprule
            \multicolumn{2}{c}{} & \multicolumn{2}{c}{\textbf{Embodied}} & \multicolumn{1}{c}{\textbf{Web}} & \multicolumn{1}{c}{\textbf{Tool-use}} \\
            \cmidrule(lr){3-4} \cmidrule(lr){5-5} \cmidrule(lr){6-6}
            \textbf{Method} & \textbf{Interface} & \textbf{ALFWorld} & \textbf{ScienceWorld} & \textbf{WebShop} & \textbf{M$^3$ToolEval} \\
            \midrule

            \multirow{2}{*}{Vanilla}
            & w/o ALIGN & 13.43 & 14.94 & 54.10 & 11.11 \\
            & w/ ALIGN & 60.45 {\scriptsize\textcolor{Green4}{(+47.02)}} & 27.69 {\scriptsize\textcolor{Green4}{(+12.75)}} & 61.23 {\scriptsize\textcolor{Green4}{(+7.13)}} & 20.83 {\scriptsize\textcolor{Green4}{(+9.72)}} \\
            \midrule

            \multirow{2}{*}{ReAct}
            & w/o ALIGN & 19.40 & 20.03 & 37.20 & \leavevmode\hphantom{0}9.72 \\
            & w/ ALIGN & 63.43 {\scriptsize\textcolor{Green4}{(+44.03)}} & 28.97 {\scriptsize\textcolor{Green4}{(+8.94)}} & 42.93 {\scriptsize\textcolor{Green4}{(+5.73)}} & 18.06 {\scriptsize\textcolor{Green4}{(+8.34)}} \\
            \midrule

            \multirow{2}{*}{Self-Consistency}
            & w/o ALIGN & 11.94 & 14.07 & 56.23 & 11.11 \\
            & w/ ALIGN & 69.40 {\scriptsize\textcolor{Green4}{(+57.46)}} & 25.41 {\scriptsize\textcolor{Green4}{(+11.34)}} & 61.10 {\scriptsize\textcolor{Green4}{(+4.87)}} & 16.67 {\scriptsize\textcolor{Green4}{(+5.56)}} \\
            \midrule

            \multirow{2}{*}{Self-Refine}
            & w/o ALIGN & \leavevmode\hphantom{0}3.73 & 14.87 & 44.80 & \leavevmode\hphantom{0}5.55 \\
            & w/ ALIGN & 40.30 {\scriptsize\textcolor{Green4}{(+36.57)}} & 22.99 {\scriptsize\textcolor{Green4}{(+8.12)}} & 52.30 {\scriptsize\textcolor{Green4}{(+7.50)}} & \leavevmode\hphantom{0}6.94 {\scriptsize\textcolor{Green4}{(+1.39)}} \\
            \midrule

            \multirow{2}{*}{Planning}
            & w/o ALIGN & \leavevmode\hphantom{0}9.70 & 17.13 & 46.95 & 11.11 \\
            & w/ ALIGN & 52.99 {\scriptsize\textcolor{Green4}{(+43.29)}} & 26.34 {\scriptsize\textcolor{Green4}{(+9.21)}} & 54.67 {\scriptsize\textcolor{Green4}{(+7.72)}} & 18.06 {\scriptsize\textcolor{Green4}{(+6.95)}} \\
            \bottomrule
        \end{tabular}
        \end{adjustbox}%
    \endgroup
    \caption{\textbf{Effect of ALIGN-generated interfaces on four benchmarks.} For every agent we report its score without the interface (w/o ALIGN) and with the interface (w/ ALIGN); the value in parentheses is the absolute improvement. Metrics are task-success rate (\%) for ALFWorld and M$^3$ToolEval, and scores for ScienceWorld and WebShop.}
    \label{tab:main_results}
\end{table*}

\section{Experiment}

\subsection{Experimental Settings\label{sec:experimental_settings}}

\textbf{Evaluation Protocol.}~
We evaluate each agent in the original environment, then re-evaluate it in the same environment wrapped with ALIGN-generated interface. Interfaces are generated only from training tasks and held fixed during testing.

\textbf{Benchmarks.}~
We use ALFWorld~\citep{DBLP:conf/iclr/ShridharYCBTH21} for embodied household tasks, ScienceWorld~\citep{DBLP:conf/emnlp/WangJCA22} for interactive scientific reasoning, WebShop~\citep{DBLP:conf/nips/Yao0YN22} for web navigation, and M$^{3}$ToolEval~\citep{DBLP:conf/icml/WangCY0L0J24} for multi-turn tool use.

\textbf{Agent Methods.}~
We evaluate Vanilla Agent (no specialized prompting), ReAct~\citep{DBLP:conf/iclr/YaoZYDSN023} (interleaved reasoning and acting), Self-Consistency~\citep{DBLP:conf/iclr/0002WSLCNCZ23} (majority voting over sampled trajectories), Self-Refine~\citep{DBLP:conf/nips/MadaanTGHGW0DPY23} (iterative self-critique) and Planning agent inspired by RAP~\cite{DBLP:conf/emnlp/HaoGMHWWH23} that decomposes tasks into sub-tasks.

\textbf{Implementation Details.}~
Unless otherwise noted, agents use Qwen2.5-7B-Instruct~\citep{qwen2.5} as the base model. Within ALIGN, the Optimizer uses Gemini\,2.5\,Pro~\citep{google2025gemini25pro} for interface code generation, and GPT-4.1~\citep{gpt4.1} is used elsewhere in the Analyzer and Optimizer; the GPT-4.1-series experiments in Section~\ref{sec:generalization} instead use the cheaper GPT-4.1-mini for these roles (Appendix~\ref{app:further}). Full setup is in Appendix~\ref{app:experiment_settings}.

\begin{table*}[htbp]
\centering
\begingroup
\fontsize{8.5}{10}\selectfont
\setlength{\tabcolsep}{4pt}
\renewcommand{\arraystretch}{1}
\begin{adjustbox}{max width=\linewidth}
\begin{tabular}{lcccccc}
\toprule
 & \multicolumn{3}{c}{ALFWorld} & \multicolumn{3}{c}{ScienceWorld}\\
\cmidrule(lr){2-4}\cmidrule(lr){5-7}
\textbf{Method} & \textbf{w/o ALIGN} & \textbf{w/ ALIGN} & $\Delta$ & \textbf{w/o ALIGN} & \textbf{w/ ALIGN} & $\Delta$\\
\midrule
Vanilla            & 77.91 & 26.59 & 66\% & 49.12 & 24.47 & 50\%\\
ReAct              & 82.23 & 38.63 & 53\% & 46.61 & 29.99 & 36\%\\
Self-Consistency   & 77.71 & 15.08 & 81\% & 51.10 & 31.51 & 38\%\\
Self-Refine        & 90.38 & 45.84 & 49\% & 58.02 & 29.48 & 49\%\\
Planning           & 74.09 & 19.14 & 74\% & 68.67 & 20.94 & 70\%\\
\midrule 
\rowcolor{blue!4}
\textbf{Average}   & 80.46 & 28.51 & 65\% & 54.70 & 27.28 & 49\%\\
\bottomrule
\end{tabular}
\end{adjustbox}
\endgroup
\caption{\textbf{Impact of the ALIGN-generated interface on consecutive invalid actions.}
The metric reports the fraction (\%) of consecutive invalid actions.
Lower values indicate more desirable behavior.
$\Delta$ denotes the relative reduction with respect to the \textbf{w/o ALIGN} setting.}
\label{tab:iface-error}
\end{table*}

\subsection{Main Results}
\label{sec:main_results}

Table~\ref{tab:main_results} summarizes the task success rates or scores of five representative agent methods in the environment without (w/o) or with (w/) ALIGN-generated interface.
The token consumption analysis can be found in Appendix~\ref{app:token}.
Our empirical investigation yields three principal findings:

\textbf{(1)~ALIGN consistently enhances performance across different domains.}
All evaluated agent methods demonstrate significant performance improvements when utilizing ALIGN-generated interfaces.
Specifically, the five agent methods exhibit mean improvements of 45.67\% in task-success rate for ALFWorld, 10.07 points for ScienceWorld, 6.59 points for WebShop, and 6.39\% in task-success rate for M$^3$ToolEval.

\textbf{(2)~Agent-environment misalignment is a pervasive phenomenon impeding the agent performance.}
The observed performance enhancements provide empirical evidence that numerous errors in baseline configurations originate from implicit constraints or under-specified observation, rather than from intrinsic reasoning deficiencies.
This finding suggests that when these environmental constraints are explicitly surfaced, agents can execute their intended tasks with substantially improved reliability.
Consequently, we posit that agent-environment misalignment is pervasive in interactive decision-making tasks, and addressing this problem is crucial for advancing agent performance.

\textbf{(3)~Alignment between agent and environment can facilitate identification of additional performance-influencing factors.}
While the Self-Consistency agent achieves a 69.40\% success rate in ALFWorld with ALIGN, the performance of Self-Refine agent remains comparatively suboptimal (40.30\%), indicating potential deficiencies in the critic and self-refinement capabilities of the Qwen2.5-7B-Instruct model.
These limitations are similarly manifested in the M$^3$ToolEval results.
Furthermore, the relatively modest performance improvements in ScienceWorld suggest that Qwen2.5-7B-Instruct may exhibit insufficient scientific causal reasoning capabilities.
These observations indicate that properly aligning agent and environment enables more precise isolation and analysis of other factors influencing agent performance beyond alignment considerations.

\subsection{Interface Quality Analysis}\label{sec:interface_quality}

\textbf{Influence on Agent Decision}~
To quantitatively assess the influence of ALIGN-generated interfaces on agent decision beyond end-task performance metric, we introduce a metric that measures the frequency of \textit{consecutive invalid actions} by calculating the proportion of the actions that occur within sequences of two or more consecutive \texttt{invalid} steps.
Lower values of this metric indicate: (1)~enhanced agent awareness of implicit preconditions, and (2)~improved recovery capability following isolated errors.
Table~\ref{tab:iface-error} presents the results for five agent methods implemented on ALFWorld and ScienceWorld.
The empirical results demonstrate a substantial reduction in consecutive invalid actions frequency across all agent methods when utilizing ALIGN-generated interfaces.
Specifically, we observe a mean reduction of 65\% in ALFWorld and 49\% in ScienceWorld.
These findings provide evidence that ALIGN effectively enriches the information conveyed by the observation, preventing agents from entering repetitive error cycles, which aligns with the findings documented in Section~\ref{sec:main_results}.

\textbf{Comparison with Agentic Systems and Human-designed Interfaces}~
To further assess the effectiveness of our automatically generated interfaces, we compare ALIGN against (1) agentic frameworks equipped with carefully designed reasoning, planning and memory modules and (2) human-designed interfaces.
The experimental setup and results are presented in Appendix~\ref{app:interface_comparison}.
As shown in Table~\ref{tab:interface_comparison_1}, even without bespoke reasoning, planning, or memory modules, a vanilla agent that directly outputs the next action yields a 6.71 percentage points higher success rate than the best agentic system when paired with ALIGN-generated interfaces, indicating agent-environment misalignment substantially constrains the performance of LLM-based agents in interactive tasks.
Moreover, using interfaces automatically generated by ALIGN yields a 13.44 percentage points higher than human-designed interfaces, further validating the effectiveness of our method~(Table~\ref{tab:interface_comparison_2}).

\subsection{Generalization and Generality Study}
\label{sec:generalization}

\textbf{Generalization Study}~
To evaluate the generalization capabilities of ALIGN, we performed the following two experiments, with the results presented in Table~\ref{tab:generalization} and detailed in Appendix~\ref{app:generalization_results}.

\begin{table}[htbp]
\small
\centering
\resizebox{\columnwidth}{!}{%
\begin{adjustbox}{max width=\linewidth}
\begin{tabular}{lrrrr}
\toprule
\rowcolor{gray!10}
\multicolumn{5}{c}{\textbf{(a) Interface source: Vanilla agent}}\\
\cmidrule{1-5}
Target method & ALF. & Sci. & Web. & M$^{3}$T.\\
\midrule
ReAct            & +39.56 & +12.29 & +7.87 & +5.56\\
Self-Consistency & +51.49 & +15.30 & +3.00 & +8.33\\
Self-Refine      & +34.33 & +14.11 & +6.17 & +4.17\\
Planning         & +41.05 &  +9.66 & +3.26 & +11.11\\
\toprule
\rowcolor{gray!10}
\multicolumn{5}{c}{\textbf{(b) Interface source: Qwen2.5-7B-Instruct agent}}\\
\cmidrule{1-5}
Target LLM & ALF. & Sci. & Web. & M$^{3}$T.\\
\midrule
Qwen2.5-14B-Instruct & +17.46 & +4.61 & +4.66 & +6.11\\
Llama3.1-8B-Instruct &  +5.97 & +10.27 & +0.33 & +0.83\\
Llama3.3-70B-Instruct & +5.82 &  +3.99 & +5.68 & +1.67\\
\bottomrule
\end{tabular}
\end{adjustbox}
}
\caption{\textbf{Generalization of ALIGN-generated interfaces across agents and models.}
Mean performance improvements from applying ALIGN-generated interfaces in the four environments across different settings. (a) Cross-agent transfer: interfaces generated with a Vanilla agent improve other agent methods. (b) Cross-model transfer: interfaces generated with Qwen2.5-7B-Instruct can generalize to other LLMs.}
\label{tab:generalization}
\end{table}

(1)~ALIGN can generalize to different agent architectures.
Panel~(a) of Table~\ref{tab:generalization} applies interfaces generated with the Vanilla agent to the other four agents.
Across all four environments every target agent shows consistent growth, demonstrating that ALIGN captures genuine and previously unexposed environment constraints.
This also reinforces the earlier conclusion that agent-environment misalignment is a pervasive source of error independent of the agent's reasoning style.

(2)~ALIGN can generalize to larger and heterogeneous LLMs.
Panel~(b) of Table~\ref{tab:generalization} examines whether an interface generated with Qwen2.5-7B-Instruct can extend to larger or architecturally different model backbones.
The results demonstrate that ALIGN-generated interfaces lead to performance improvements across base models of varying sizes and architectural families, which indicates that our method possesses strong generalization capabilities.
We also observe that this generalization is not uniformly robust across all model families and datasets.
For instance, Llama3.1-8B-Instruct~\citep{llama3.1} shows only a marginal gain of +0.33 on the WebShop benchmark.
This limited improvement may be attributed to the inherent reasoning capabilities of the model itself.

Taken together, these results show that ALIGN-generated interfaces can generalize (1) across agent policies and (2) across model scales and families, validating the practicality of ALIGN.

\begin{table}[htbp]
\small
\centering
\begin{adjustbox}{max width=\linewidth}
\begin{tabular}{llc}
\toprule
\rowcolor{gray!10}
\multicolumn{3}{c}{\textbf{(a) GPT-4.1 series}}\\
      \cmidrule{1-3}
      Base Model & Interface & SR~(\%) \\
      \midrule
      \multirow{2}{*}{GPT-4.1-mini}
      & w/o ALIGN & 28.36 \\
      & w/ ALIGN  & 64.93 {\scriptsize\textcolor{Green4}{(+36.57)}} \\
      \midrule
      \multirow{2}{*}{GPT-4.1}
      & w/o ALIGN & 73.88 \\
      & w/ ALIGN  & 93.28 {\scriptsize\textcolor{Green4}{(+19.40)}} \\
      \toprule
\rowcolor{gray!10}
\multicolumn{3}{c}{\textbf{(b) GiGPO-Qwen2.5-7B-Instruct}}\\
      \cmidrule{1-3}
      Agent Method & Interface & SR~(\%) \\
      \midrule
      \multirow{2}{*}{Vanilla}
      & w/o ALIGN & 35.04 \\
      & w/ ALIGN  & 55.97 {\scriptsize\textcolor{Green4}{(+20.93)}} \\
      \midrule
      \multirow{2}{*}{Training Config}
      & w/o ALIGN & 89.55 \\
      & w/ ALIGN  & 92.54 {\scriptsize\textcolor{Green4}{(+2.99)}}\leavevmode\hphantom{0} \\
      \bottomrule
\end{tabular}
\end{adjustbox}
\caption{\textbf{Generality of ALIGN.}
Task success rates~(SR) without and with ALIGN-generated interfaces in ALFWorld across two settings. (a) Using GPT-4.1 series models as the base model; (b) Using GiGPO-Qwen2.5-7B-Instruct under different agent architectures.}
\label{tab:generality}
\end{table}

\textbf{Generality Study}~
In this work, our empirical observations indicate that the root cause of agent-environment misalignment lies in the robustness of the interface itself, making it a universal issue that affects agents irrespective of the underlying model capability.
To further validate this claim and assess the generality of ALIGN, we conduct experiments on both closed-source LLMs and domain-specific models trained within the environment.
For the former, we use the GPT-4.1 series; for the latter, we use GiGPO-Qwen2.5-7B-Instruct-ALFWorld~\citep{DBLP:journals/corr/abs-2505-10978}, a state-of-the-art model specifically post-trained on ALFWorld via reinforcement learning.
Detailed experimental setup and full results are provided in Appendix~\ref{app:further}.
As the results reported in Panel~(a) of Table~\ref{tab:generality} shown, applying the ALIGN-generated interface substantially improves the performance of the GPT-4.1-based agent from 73.88\% to 93.28\%.
Meanwhile, as the results reported in Panel~(b) of Table~\ref{tab:generality} shown, the ALIGN-generated interface also enhances the performance of the domain-specific model under both our Vanilla Agent setting and its original training configuration, from 35.04\% to 55.97\% and 89.55\% to 92.54\%, respectively.
These findings demonstrate that the fundamental and pervasive nature of agent-environment misalignment stems from deficiencies in the environment’s interface rather than solely from the reasoning limitations of any given model, and further corroborate the generality of our method across both frontier and domain-specialized models.

\subsection{Ablation Study}

\textbf{Ablation on Interface Components}~
Starting from the full ALIGN interface, we ablate \textsc{InferRules} and \textsc{WrapStep} separately.
As shown in Table~\ref{tab:ablation_interface}, both ablations reduce performance: w/o \textsc{InferRules} averages -6.72 percentage points on ALFWorld and -2.05 on ScienceWorld, while removing \textsc{WrapStep} yields a larger decline of -31.79 percentage points and -7.84, respectively.
These decreases confirm that each interface component contributes meaningfully.
Moreover, the much larger drop w/o \textsc{WrapStep} shows the critical role of fine-grained and enriched observation during interaction.
This also suggests that rich, LLM-friendly observation should be prioritized by future environment designers when constructing environments.

\begin{table}[t]
\small
\centering
\setlength{\tabcolsep}{3pt}
\begin{adjustbox}{max width=\linewidth}
\begin{tabular}{lcccc}
\toprule
& \multicolumn{2}{c}{\textbf{w/o \textsc{InferRules}}} & \multicolumn{2}{c}{\textbf{w/o \textsc{WrapStep}}}\\
\cmidrule(lr){2-3}\cmidrule(lr){4-5}
Method & ALF. & Sci. & ALF. & Sci.\\
\midrule
Vanilla           & \leavevmode\hphantom{0}-8.96 & -3.35 & -33.58 & \leavevmode\hphantom{0}-4.72 \\
ReAct             &  \leavevmode\hphantom{0}-5.22 & -2.08 & -17.91 & \leavevmode\hphantom{0}-6.44 \\
Self-Consistency  &  \leavevmode\hphantom{0}-1.49 & -2.30 & -46.27 & -10.59\\
Self-Refine       &  \leavevmode\hphantom{0}-7.46 & -1.72 & -34.33 & \leavevmode\hphantom{0}-7.59 \\
Planning          &  -10.45 & -0.78 & -26.87 & \leavevmode\hphantom{0}-9.86 \\
\midrule
\rowcolor{blue!4}
\textit{Mean}     & \leavevmode\hphantom{0}-6.72  & -2.05 & -31.79 & \leavevmode\hphantom{0}-7.84 \\
\bottomrule
\end{tabular}
\end{adjustbox}
\caption{\textbf{Ablation on Interface components.}
Values show performance change on ALFWorld and ScienceWorld.
Negative values mean drops from \emph{Full} interface.
}
\label{tab:ablation_interface}
\end{table}

\begin{table}[htbp]
  \footnotesize
  \centering
  \begin{adjustbox}{max width=\linewidth}
  \begin{tabular}{@{}ccccc@{}}
    \toprule
  \textbf{Temp.}&\textbf{Iter0}&\textbf{Iter1}&\textbf{Iter2}&\textbf{Iter3}\\\midrule
    0.2 & 13.43 & 22.39 & 0.00 & 0.00\\
    0.5 & 13.43 & 23.88 & 1.49 & 0.75\\
  \bottomrule
  \end{tabular}
  \end{adjustbox}
  \caption{Task success rate~(\%) on ALFWorld across iterations without experimental verification procedure.}
  \label{tab:sim-ablation}
\end{table}

\textbf{Ablation on Experimental Verification}~
To assess whether the experimental verification procedure in Section~\ref{sec:ALIGN_framework} is indispensable, we ablated it with the Vanilla agent on ALFWorld.
As a surrogate, the Analyzer and Optimizer each sampled six candidates per iteration and selected their top one, under decoding temperatures $T\in\{0.2,0.5\}$.
As shown in Table~\ref{tab:sim-ablation}, success rate deteriorates sharply across iterations, reflecting the limited single-shot reliability of LLMs in diagnosing misalignments and synthesizing interfaces, and confirming the necessity of experimental verification.

\subsection{Further Analyses}
\label{sec:further_analyses}

\begin{table}[t]
\small
\centering
\setlength{\tabcolsep}{4pt}
\resizebox{\columnwidth}{!}{%
\begin{adjustbox}{max width=\linewidth}
\begin{tabular}{lcccc}
\toprule
\textbf{Base Model} & \textbf{Baseline} & \textbf{+AutoManual} & \textbf{+ALIGN} & \textbf{+Both}\\
\midrule
Qwen2.5-7B-Instruct & 13.43 & 29.10 & 60.45 & \textbf{64.93}\\
Qwen3-8B            & 15.67 & 52.24 & 47.76 & \textbf{53.73}\\
\bottomrule
\end{tabular}
\end{adjustbox}%
}
\caption{\textbf{Complementarity of ALIGN with agent-side rule learning.} Task success rates~(\%) of the Vanilla agent on ALFWorld with AutoManual's released rules, the ALIGN-generated interface, and both; Qwen3-8B reuses the Qwen2.5-7B-Instruct interface.}
\label{tab:complementarity}
\end{table}

\textbf{Complementarity with Agent-Side Rule Learning}~
We test the complementarity claimed in Section~\ref{sec:related} by injecting the ten rules released by AutoManual~\citep{DBLP:conf/nips/0001LYYL024} into the agent prompt, exactly as its testing stage consumes a manual (adaptation details in Appendix~\ref{app:complementarity}).
Table~\ref{tab:complementarity} reports results for Qwen2.5-7B-Instruct and Qwen3-8B~\citep{qwen3}, the latter reusing the Qwen2.5-7B-Instruct interface (cf.\ Table~\ref{tab:generalization}(b)).
The combination is the best configuration for both models, and the two methods repair different levels of failure: AutoManual's rules operate at the plan level (task-type recipes), whereas ALIGN operates at the execution level (why a step failed and which precondition applies).

\textbf{Content Audit of Injected Observations}~
Interfaces are generated from training tasks only and frozen at test time, so the remaining question is one of content.
We audited all 312 agent-facing strings injected by the 20 final interfaces (Appendix~\ref{app:content_audit}): an injection is \emph{mechanic-level} iff it is a deterministic function of the current state and the attempted action and contains no task-goal content.
ALFWorld and WebShop contain no task-specific guidance, so the headline results rest on mechanic-level content only; a small minority in ScienceWorld and M$^3$ToolEval (21 of 312 strings) encodes procedure-level knowledge, which we trace to their one- or two-task training splits and mitigate with a task-agnosticism constraint in the Optimizer prompt (Appendix~\ref{app:prompt}).

\textbf{Exploration and Strategy Diversity}~
Analyzing the complete evaluation trajectories (Appendix~\ref{app:behavior}), we find that action-type entropy stays in the same band with and without ALIGN, that agents visit more distinct locations per task with ALIGN, and that the gains consist of failure-to-success conversions (50--79 tasks per method on ALFWorld) against only 1--7 regressions: the signature of restored observability rather than lowered difficulty.

\textbf{Generator Capability and Cost}~
Generation is a one-time cost per environment: the full $K{=}8$ iterations on ALFWorld cost about USD~9.4 at list prices, i.e., about USD~0.003 per task-run over the reuse in this paper, while the interface itself involves no model calls and reduces per-task inference to 58\% of the baseline's input tokens and 43\% of its LLM calls (Appendix~\ref{app:cost}).
Only the Optimizer's code synthesis requires a frontier model: with GPT-4.1-mini as the Analyzer and verification model, the generated interface is no worse than that of the frontier stack (64.93\% vs.\ 60.45\% for Qwen2.5-7B-Instruct), whereas a fully self-sufficient Qwen2.5-7B-Instruct stack fails to complete generation (Appendix~\ref{app:generator_spectrum}); the agent itself has no such floor.

\section{Conclusion}

In this work, we introduce \textbf{ALIGN}, a framework that automatically generates aligned interfaces to alleviate the \textbf{agent-environment misalignment} in interactive decision-making tasks.
By diagnosing implicit constraints via the Analyzer and synthesizing interfaces via the Optimizer, ALIGN achieves significant improvements across four benchmarks spanning embodied tasks, web navigation, and tool-use.
Our results show that ALIGN generalizes to unseen models and strategies, offering a robust, plug-and-play solution that decouples agent design from manual environment-specific alignment.

\section*{Limitations\label{app:limitations}}
Despite the effectiveness of ALIGN in alleviating agent-environment misalignment, this work represents an initial exploration into automated interface generation, and several boundaries should be stated plainly.

\textbf{Scope: text-based environments with a programmable boundary.}~
All of our evaluation is conducted in text-based environments with discrete action spaces, spanning embodied tasks, web navigation, and tool use.
ALIGN applies wherever the agent--system boundary is programmable, which holds for tool-use, API, web, and code agents, whose production systems are already mediated by hand-built wrapper layers such as tool schemas and agent-computer interfaces~\citep{DBLP:conf/nips/YangJWLYNP24}; physical robots and raw pixel streams fall outside this condition.
Although the formalization in Section~\ref{sec:3.1} is modality-agnostic, and the textual projections perceived by current GUI agents (accessibility trees and Set-of-Marks element lists~\citep{DBLP:journals/corr/abs-2310-11441}) would allow \textsc{InferRules} and \textsc{WrapStep} to apply as they are, we claim no GUI results: pixel-level observation enrichment remains an open problem, and the accessibility-tree instantiation is the natural next step.

\textbf{Supervision and undiagnosable causes.}~
The Analyzer requires the environment's success/failure signal and the agent's own failed trajectories, but no human labels or gold trajectories; settings with no outcome signal at all are not covered.
Moreover, every hypothesized misalignment must be experimentally confirmed before an interface change is accepted, so a cause that the Analyzer cannot verify (e.g., sensor noise or undocumented API behavior) is discarded and the interface simply remains unchanged: the failure mode is graceful, but no improvement is obtained.

\textbf{Over-guidance and cross-family robustness.}~
The content audit in Appendix~\ref{app:content_audit} finds a small amount of procedure-level content in the two domains with one- or two-task training splits; we mitigate this with a task-agnosticism constraint in the Optimizer, but larger training splits would be a more principled remedy.
In addition, an interface generated with one model family may not recognize the action surface forms of another (e.g., Qwen3-8B emits ``use desklamp 1 on X'' where Qwen2.5 emits ``use desklamp 1''), so generating family-general interfaces is a direction for future work.

\textbf{Metrics for interface quality.}~
Beyond task success rates and consecutive invalid actions, we report behavioral diagnostics on action diversity and exploration in Appendix~\ref{app:behavior}.
More comprehensive metrics, such as LLM-as-a-Judge~\citep{DBLP:conf/nips/ZhengC00WZL0LXZ23} evaluation of whether an interface conveys environment constraints, remain a promising direction.

\section*{Acknowledgments}

This work is supported by Fundamental and Interdisciplinary Disciplines Breakthrough Plan of the Ministry of Education of China (No. JYB2025XDXM101), the National Natural Science Foundation of China (No. 62276152, 62236011), funding from Wuxi Research Institute of Applied Technologies, Tsinghua University under Grant 20242001120, and Key Laboratory of Ethnic Language Intelligent Analysis and Security Governance of MOE, Minzu University of China, Beijing, China.

\bibliographystyle{acl_natbib}
\bibliography{custom}

\flushcolsend
\clearpage
\raggedcolsend

\appendix

\section{LLM Usage Statement}

Throughout the completion of this work, the LLM was employed solely for the purpose of refining
sentences and improving grammatical accuracy during the manuscript writing process.

\section{Formalization of the ALIGN Algorithm\label{app:formalization}}

The formalization of the ALIGN algorithm is outlined in Algorithm~\ref{alg:align}.

\begin{algorithm}[htbp]
\small
\caption{\small ALIGN: Auto-Aligned Interface Generation}
\label{alg:align}
\begin{algorithmic}[1]
\Require{Environment $\mathcal{E}$, Agent $\pi$, Task training set $\mathcal{T}_{\text{train}}$, Maximum iterations $K$}
\State Initialize misalignment set $\mathcal{M} \gets \emptyset$, interface $\Phi^{(0)} \gets \{\textsc{InferRules}^{(0)}, \textsc{WrapStep}^{(0)}\}$, where $\textsc{InferRules}^{(0)}$ and $\textsc{WrapStep}^{(0)}$ are identity functions
\For{$i = 1,2,\dots,K$}
    \State $\tilde{\mathcal{E}}^{(i-1)} \gets$ Environment $\mathcal{E}$ wrapped with interface $\Phi^{(i-1)}$
    \State $\tau_{\text{fail}}^{(i-1)} \gets$ Failed trajectories from agent $\pi$ interacting with $\tilde{\mathcal{E}}^{(i-1)}$ on $\mathcal{T}_{\text{train}}$
    \If{$\tau_{\text{fail}}^{(i-1)} = \emptyset$}
        \State \textbf{break} \Comment{No more failures in the training set}
    \EndIf

    \Statex\hspace{\algorithmicindent}\textcolor{gray}{//\,Stage 1: Misalignment Analysis}
    \State $\mathcal{M}^{(i)} \gets \text{Analyzer}(\tau_{\text{fail}}^{(i-1)}, \mathcal{M}, \Phi^{(i-1)})$
    \If{$\mathcal{M}^{(i)} = \emptyset$}
        \State \textbf{break} \Comment{No new misalignments identified}
    \EndIf
    \State $\mathcal{M} \gets \mathcal{M} \cup \mathcal{M}^{(i)}$
    
    \Statex\hspace{\algorithmicindent}\textcolor{gray}{//\,Stage 2: Interface Generation}
    \State $\Phi^{(i)} \gets \text{Optimizer}(\mathcal{M}^{(i)}, \Phi^{(i-1)})$
\EndFor
\State \textbf{return} final interface $\Phi^{(i)}$
\end{algorithmic}
\end{algorithm}

\section{Supplementary Experimental Setup and Detailed Results}
\label{app:supp}

\subsection{Preliminary Experiments\label{app:preliminary}}

To preliminarily assess the significance of agent-environment misalignment, we conducted exploratory experiments on the ALFWorld.
We employed the vanilla Qwen2.5-7B-Instruct agent with a temperature setting of 0.0.
The deployment protocol, prompt template, followed the same configuration described in Appendix~\ref{app:experiment_settings} and Appendix~\ref{app:prompt}.

During the experiments, we introduced a minor modification to the environment: if the agent issued the action \textit{examine receptacle} and the environment returned the default observation ``Nothing happens.'', we replaced it with ``You need to first go to receptacle before you can examine it.''
This simple adjustment increased the agent's task success rate from 13.4\% to 31.3\%.

\subsection{Interface Quality Analysis Experiments\label{app:interface_comparison}}

\begin{table*}[htbp]
\centering
\resizebox{\textwidth}{!}{%
\begin{adjustbox}{max width=\linewidth}
\begin{tabular}{lllccccccc}
\toprule
\multicolumn{1}{c}{\textbf{\makecell{Agent\\Framework}}} &
\textbf{Interface} &
\multicolumn{1}{c}{\textbf{\makecell{Memory\\Module}}} &
\multicolumn{1}{c}{\textbf{\makecell{pick\\and place}}} &
\multicolumn{1}{c}{\textbf{\makecell{pick clean\\and place}}} &
\multicolumn{1}{c}{\textbf{\makecell{pick heat\\and place}}} &
\multicolumn{1}{c}{\textbf{\makecell{pick cool\\and place}}} &
\multicolumn{1}{c}{\textbf{\makecell{look at /\\examine\\in light}}} &
\multicolumn{1}{c}{\textbf{\makecell{pick two obj\\and place}}} &
\multicolumn{1}{c}{\textbf{\makecell{Success\\Rate (\%)}}} \\
\midrule
AgentSquare & / & Generative & \leavevmode\hphantom{0}95.83 & \leavevmode\hphantom{0}87.10 & 69.57 & \leavevmode\hphantom{0}95.24 & 83.33 & \leavevmode\hphantom{0}88.24 & 86.57 \\
\midrule
AgentSquare & / & DiLu & \leavevmode\hphantom{0}91.67 & \leavevmode\hphantom{0}87.10 & 52.17 & \leavevmode\hphantom{0}95.24 & 83.33 & \leavevmode\hphantom{0}70.59 & 80.60 \\
\midrule
AgentSquare & / & TP & \leavevmode\hphantom{0}87.50 & \leavevmode\hphantom{0}51.61 &  \leavevmode\hphantom{0}4.35 & \leavevmode\hphantom{0}61.90 & 27.78 & \leavevmode\hphantom{0}47.06 & 47.76 \\
\midrule
AgentSquare & / & VOYAGER & \leavevmode\hphantom{0}95.83 & \leavevmode\hphantom{0}83.87 & 52.17 & \leavevmode\hphantom{0}90.48 & 83.33 & \leavevmode\hphantom{0}64.71 & 79.10 \\
\midrule
Vanilla Agent & w/o ALIGN & / & 100.00 & \leavevmode\hphantom{0}93.55 & 13.04 & \leavevmode\hphantom{0}71.43 & 61.11 & 100.00 & 73.88 \\
\midrule
Vanilla Agent & w/ ALIGN & / & 100.00 & 100.00 & 78.26 & 100.00 & 77.78 & 100.00 & \textbf{93.28} \\
\bottomrule
\end{tabular}
\end{adjustbox}
}
\caption{Experimental results of the comparison between agents with ALIGN-generated interface and agents with human-designed reasoning, planning and memory module.}
\label{tab:interface_comparison_1}
\end{table*}

To further assess the quality of the ALIGN-generated interface, we first compare our method with human-designed agentic system.
Our experiments are conducted on ALFWorld using the AgentSquare~\citep{DBLP:conf/iclr/ShangLZMLXL25} framework.
To maximize the advantages of the agentic system, we adopt gpt-4.1-2025-04-14 as the base model, select OPENAGI~\citep{DBLP:conf/nips/GeHMJTXLZ23} for the planning module, Self-Refine~\citep{DBLP:conf/nips/MadaanTGHGW0DPY23} for the reasoning module, and evaluate memory using Generative~\citep{DBLP:conf/uist/ParkOCMLB23}, DiLu~\citep{DBLP:conf/iclr/WenF0C0CDS0024}, TP~\citep{DBLP:conf/iclr/Yu0Y24}, and VOYAGER~\citep{DBLP:journals/tmlr/WangX0MXZFA24}.
For our approach, we employ a gpt-4.1-2025-04-14-based vanilla agent, where the interface is generated with the gpt-4.1-mini-2025-04-14-based vanilla agent by ALIGN (the experimental setup is same as Appendix~\ref{app:further}).
The results are reported in Table~\ref{tab:interface_comparison_1}.

\begin{table}[htbp]
    \footnotesize
    \centering
    \resizebox{.38\textwidth}{!}{%
    \begin{adjustbox}{max width=\linewidth}
    \begin{tabular}{@{}lc@{}}
        \toprule
        \textbf{Experimental Setting} & \textbf{Success Rate (\%)} \\
        \midrule
        w/o Interface & 13.43 \\
        \midrule
        Few-shot  & 44.78 \\
        \midrule
        Valid Actions  & 44.03 \\
        \midrule
        Human Designed Interface  & 47.01 \\
        \midrule
        ALIGN-generated Interface  & \textbf{60.45} \\
        \bottomrule
    \end{tabular}
    \end{adjustbox}%
    }
    \caption{Experimental results of the comparison between agents with ALIGN-generated interface and agents with human-designed interfaces.}
    \label{tab:interface_comparison_2}
\end{table}

Furthermore, we compare the ALIGN-generated interface against the human-designed interface.
We adopt the following configurations for comparison with our method:
(1)~\textbf{Few-shot:} Settings identical to those in the ReAct~\citep{DBLP:conf/iclr/YaoZYDSN023}; (2) \textbf{Valid Actions:} Supplying the agent with all valid actions at every response turn, analogous to the \texttt{check\_valid\_actions} configuration in AgentBoard~\citep{DBLP:conf/nips/MaZZYYJLKH24}; (3) \textbf{Human-Designed Interface:} Interfaces manually crafted by Ph.D. students after inspecting ALFWorld experiments, examining trajectories, and running experiments themselves. The design logic includes: executing ``go to'' prior to each action; automatically checking object labels; converting ``put'' to ``move'' when appropriate; returning the action space upon invalid actions; issuing reminders when ``clean with'' is applied to non-sinkbasin objects; and other hand-engineered rules.
We use Qwen2.5-7B-Instruct as the base model.
Experimental results are reported in Table~\ref{tab:interface_comparison_2}.

\subsection{Generalization Study Experiments\label{app:generalization_results}}

Detailed results of the generalization study are provided for the cross-method experiments in Table~\ref{tab:full_results_generalization_method} and for the cross-model experiments in Tables~\ref{tab:full_results_generalization_model_qwen14b}, \ref{tab:full_results_generalization_model_llama3_8b}, and \ref{tab:full_results_generalization_model_llama3_70b}.

\begin{table*}[htbp]
    \small
    \centering
        \begin{adjustbox}{max width=\linewidth}
        \begin{tabular}{@{}llcccc@{}}
            \toprule
            \multicolumn{2}{c}{\textbf{Base Method: Vanilla}} & \multicolumn{2}{c}{\textbf{Embodied}} & \textbf{Web} & \textbf{Tool-use} \\
            \cmidrule(lr){3-4} \cmidrule(lr){5-5} \cmidrule(lr){6-6}
            \textbf{Method} & \textbf{Interface} & \textbf{ALFWorld} & \textbf{ScienceWorld} & \textbf{WebShop} & \textbf{M$^{3}$ToolEval} \\
            \midrule
            \multirow{2}{*}{ReAct}
              & w/o ALIGN & 19.40 & 20.03 & 37.20 & 9.72 \\
              & w/ ALIGN  & 58.96 {\scriptsize\textcolor{Green4}{(+39.56)}} &
                             32.32 {\scriptsize\textcolor{Green4}{(+12.29)}} &
                             45.07 {\scriptsize\textcolor{Green4}{(+7.87)}} &
                             15.28 {\scriptsize\textcolor{Green4}{(+5.56)}} \\
            \midrule
            \multirow{2}{*}{Self-Consistency}
              & w/o ALIGN & 11.94 & 14.07 & 56.23 & 11.11 \\
              & w/ ALIGN  & 63.43 {\scriptsize\textcolor{Green4}{(+51.49)}} &
                             29.37 {\scriptsize\textcolor{Green4}{(+15.30)}} &
                             59.23 {\scriptsize\textcolor{Green4}{(+3.00)}} &
                             19.44 {\scriptsize\textcolor{Green4}{(+8.33)}} \\
            \midrule
            \multirow{2}{*}{Self-Refine}
              & w/o ALIGN & 3.73  & 14.87 & 44.80 & 5.55 \\
              & w/ ALIGN  & 38.06 {\scriptsize\textcolor{Green4}{(+34.33)}} &
                             28.98 {\scriptsize\textcolor{Green4}{(+14.11)}} &
                             50.97 {\scriptsize\textcolor{Green4}{(+6.17)}} &
                             9.72  {\scriptsize\textcolor{Green4}{(+4.17)}} \\
            \midrule
            \multirow{2}{*}{Planning}
              & w/o ALIGN & 9.70  & 17.13 & 46.95 & 11.11 \\
              & w/ ALIGN  & 50.75 {\scriptsize\textcolor{Green4}{(+41.05)}} &
                             26.79 {\scriptsize\textcolor{Green4}{(+9.66)}}  &
                             50.21 {\scriptsize\textcolor{Green4}{(+3.26)}} &
                             22.22 {\scriptsize\textcolor{Green4}{(+11.11)}} \\
            \bottomrule
        \end{tabular}
        \end{adjustbox}%
    \caption{\textbf{Generalization of ALIGN-generated interfaces generated with Vanilla agents to other agent methods.}
    For each agent we report its score without the interface (w/o~ALIGN) and with the interface (w/~ALIGN); the value in parentheses is the \emph{absolute} improvement.}
    \label{tab:full_results_generalization_method}
\end{table*}

\begin{table*}[htbp]
    \small
    \centering
        \begin{adjustbox}{max width=\linewidth}
        \begin{tabular}{@{}llcccc@{}}
            \toprule
            \multicolumn{2}{c}{\textbf{Base Model: Qwen2.5-14B-Instruct}}
              & \multicolumn{2}{c}{\textbf{Embodied}} & \textbf{Web} & \textbf{Tool-use} \\
            \cmidrule(lr){3-4} \cmidrule(lr){5-5} \cmidrule(lr){6-6}
            \textbf{Method} & \textbf{Interface} & \textbf{ALFWorld} & \textbf{ScienceWorld} & \textbf{WebShop} & \textbf{M$^{3}$ToolEval} \\
            \midrule
            \multirow{2}{*}{Vanilla}
              & w/o ALIGN & 48.51 & 22.58 & 53.67 & 13.89 \\
              & w/ ALIGN  & 52.24 {\scriptsize\textcolor{Green4}{(+3.73)}} &
                             37.58 {\scriptsize\textcolor{Green4}{(+15.00)}} &
                             58.40 {\scriptsize\textcolor{Green4}{(+4.73)}} &
                             18.06 {\scriptsize\textcolor{Green4}{(+4.17)}} \\
            \midrule
            \multirow{2}{*}{ReAct}
              & w/o ALIGN & 54.48 & 31.24 & 39.73 & 15.28 \\
              & w/ ALIGN  & 70.15 {\scriptsize\textcolor{Green4}{(+15.67)}} &
                             29.79 {\scriptsize\textcolor{Red}{(-1.45)}} &
                             42.17 {\scriptsize\textcolor{Green4}{(+2.44)}} &
                             26.39 {\scriptsize\textcolor{Green4}{(+11.11)}} \\
            \midrule
            \multirow{2}{*}{Self-Consistency}
              & w/o ALIGN & 43.28 & 25.60 & 52.63 & 13.89 \\
              & w/ ALIGN  & 72.39 {\scriptsize\textcolor{Green4}{(+29.11)}} &
                             26.68 {\scriptsize\textcolor{Green4}{(+1.08)}} &
                             51.07 {\scriptsize\textcolor{Red}{(-1.56)}} &
                             27.78 {\scriptsize\textcolor{Green4}{(+13.89)}} \\
            \midrule
            \multirow{2}{*}{Self-Refine}
              & w/o ALIGN & 5.22  & 18.97 & 41.00 & 15.28 \\
              & w/ ALIGN  & 14.18 {\scriptsize\textcolor{Green4}{(+8.96)}} &
                             20.72 {\scriptsize\textcolor{Green4}{(+1.75)}} &
                             39.93 {\scriptsize\textcolor{Red}{(-1.07)}} &
                             16.67 {\scriptsize\textcolor{Green4}{(+1.39)}} \\
            \midrule
            \multirow{2}{*}{Planning}
              & w/o ALIGN & 49.25 & 21.46 & 31.72 & 25.00 \\
              & w/ ALIGN  & 79.10 {\scriptsize\textcolor{Green4}{(+29.85)}} &
                             28.13 {\scriptsize\textcolor{Green4}{(+6.67)}} &
                             50.47 {\scriptsize\textcolor{Green4}{(+18.75)}} &
                             25.00 {\scriptsize(0.00)} \\
            \bottomrule
        \end{tabular}
        \end{adjustbox}%
    \caption{\textbf{Generalization of ALIGN-generated interfaces generated with Qwen2.5-7B-Instruct to Qwen2.5-14B-Instruct.}
    For each agent we report its score without the interface (w/o~ALIGN) and with the interface (w/~ALIGN); the value in parentheses is the \emph{absolute} improvement.}
    \label{tab:full_results_generalization_model_qwen14b}
\end{table*}

\begin{table*}[htbp]
    \small
    \centering
        \begin{adjustbox}{max width=\linewidth}
        \begin{tabular}{@{}llcccc@{}}
            \toprule
            \multicolumn{2}{c}{\textbf{Base Model: Llama3.1-8B-Instruct}}
              & \multicolumn{2}{c}{\textbf{Embodied}} & \textbf{Web} & \textbf{Tool-use} \\
            \cmidrule(lr){3-4} \cmidrule(lr){5-5} \cmidrule(lr){6-6}
            \textbf{Method} & \textbf{Interface} & \textbf{ALFWorld} & \textbf{ScienceWorld} & \textbf{WebShop} & \textbf{M$^{3}$ToolEval} \\
            \midrule
            \multirow{2}{*}{Vanilla}
              & w/o ALIGN & 5.22 & 23.59 & 35.17 & 5.56 \\
              & w/ ALIGN  & 14.18 {\scriptsize\textcolor{Green4}{(+8.96)}} &
                             36.40 {\scriptsize\textcolor{Green4}{(+12.81)}} &
                             24.00 {\scriptsize\textcolor{Red}{(-11.17)}}  &
                             1.39 {\scriptsize\textcolor{Red}{(-4.17)}} \\
            \midrule
            \multirow{2}{*}{ReAct}
              & w/o ALIGN & 1.49 & 22.42 & 27.12 & 12.50 \\
              & w/ ALIGN  & 15.67 {\scriptsize\textcolor{Green4}{(+14.18)}} &
                             28.74 {\scriptsize\textcolor{Green4}{(+6.32)}} &
                             27.10 {\scriptsize\textcolor{Red}{(-0.02)}}                   &
                             22.22 {\scriptsize\textcolor{Green4}{(+9.72)}} \\
            \midrule
            \multirow{2}{*}{Self-Consistency}
              & w/o ALIGN & 5.22 & 25.21 & 29.80 & 4.17 \\
              & w/ ALIGN  & 11.94 {\scriptsize\textcolor{Green4}{(+6.72)}} &
                             34.83 {\scriptsize\textcolor{Green4}{(+9.62)}} &
                             15.83 {\scriptsize\textcolor{Red}{(-13.97)}}  &
                             2.78 {\scriptsize\textcolor{Red}{(-1.39)}} \\
            \midrule
            \multirow{2}{*}{Self-Refine}
              & w/o ALIGN & 0.00       & 22.34 & 27.70 & 1.39 \\
              & w/ ALIGN  & 0.75 {\scriptsize\textcolor{Green4}{(+0.75)}} &
                             31.33 {\scriptsize\textcolor{Green4}{(+8.99)}} &
                             37.43 {\scriptsize\textcolor{Green4}{(+9.73)}} &
                             1.39 {\scriptsize(0.00)} \\
            \midrule
            \multirow{2}{*}{Planning}
              & w/o ALIGN & 6.72 & 13.33 & 23.67 & 4.17 \\
              & w/ ALIGN  & 5.97 {\scriptsize\textcolor{Red}{(-0.75)}}  &
                             26.95 {\scriptsize\textcolor{Green4}{(+13.62)}} &
                             40.77 {\scriptsize\textcolor{Green4}{(+17.10)}} &
                             4.17 {\scriptsize(0.00)} \\
            \bottomrule
        \end{tabular}
        \end{adjustbox}%
    \caption{\textbf{Generalization of ALIGN-generated interfaces generated with Qwen2.5-7B-Instruct to Llama3.1-8B-Instruct.}
    For each agent we report its score without the interface (w/o~ALIGN) and with the interface (w/~ALIGN); the value in parentheses is the \emph{absolute} improvement.}
    \label{tab:full_results_generalization_model_llama3_8b}
\end{table*}

\begin{table*}[htbp]
    \small
    \centering
        \begin{adjustbox}{max width=\linewidth}
        \begin{tabular}{@{}llcccc@{}}
            \toprule
            \multicolumn{2}{c}{\textbf{Base Model: Llama3.3-70B-Instruct}}
              & \multicolumn{2}{c}{\textbf{Embodied}} & \textbf{Web} & \textbf{Tool-use} \\
            \cmidrule(lr){3-4} \cmidrule(lr){5-5} \cmidrule(lr){6-6}
            \textbf{Method} & \textbf{Interface} & \textbf{ALFWorld} & \textbf{ScienceWorld} & \textbf{WebShop} & \textbf{M$^{3}$ToolEval} \\
            \midrule
            \multirow{2}{*}{Vanilla}
              & w/o ALIGN & 52.99 & 55.77 & 51.67 & 37.50 \\
              & w/ ALIGN  & 43.28 {\scriptsize\textcolor{Red}{(-9.71)}} &
                             57.74 {\scriptsize\textcolor{Green4}{(+1.97)}} &
                             62.07 {\scriptsize\textcolor{Green4}{(+10.40)}} &
                             33.33 {\scriptsize\textcolor{Red}{(-4.17)}} \\
            \midrule
            \multirow{2}{*}{ReAct}
              & w/o ALIGN & 45.52 & 56.50 & 58.22 & 34.72 \\
              & w/ ALIGN  & 47.01 {\scriptsize\textcolor{Green4}{(+1.49)}} &
                             58.28 {\scriptsize\textcolor{Green4}{(+1.78)}} &
                             53.83 {\scriptsize\textcolor{Red}{(-4.39)}} &
                             43.06 {\scriptsize\textcolor{Green4}{(+8.34)}} \\
            \midrule
            \multirow{2}{*}{Self-Consistency}
              & w/o ALIGN & 54.48 & 56.66 & 50.37 & 36.11 \\
              & w/ ALIGN  & 65.67 {\scriptsize\textcolor{Green4}{(+11.19)}} &
                             59.24 {\scriptsize\textcolor{Green4}{(+2.58)}} &
                             55.63 {\scriptsize\textcolor{Green4}{(+5.26)}} &
                             34.72 {\scriptsize\textcolor{Red}{(-1.39)}} \\
            \midrule
            \multirow{2}{*}{Self-Refine}
              & w/o ALIGN & 38.06 & 56.97 & 38.40 & 1.39 \\
              & w/ ALIGN  & 46.27 {\scriptsize\textcolor{Green4}{(+8.21)}} &
                             60.17 {\scriptsize\textcolor{Green4}{(+3.20)}} &
                             47.85 {\scriptsize\textcolor{Green4}{(+9.45)}} &
                             0.00 {\scriptsize\textcolor{Red}{(-1.39)}} \\
            \midrule
            \multirow{2}{*}{Planning}
              & w/o ALIGN & 58.96 & 48.75 & 54.90 & 33.33 \\
              & w/ ALIGN  & 76.87 {\scriptsize\textcolor{Green4}{(+17.91)}} &
                             59.17 {\scriptsize\textcolor{Green4}{(+10.42)}} &
                             62.60 {\scriptsize\textcolor{Green4}{(+7.70)}} &
                             40.28 {\scriptsize\textcolor{Green4}{(+6.95)}} \\
            \bottomrule
        \end{tabular}
        \end{adjustbox}%
    \caption{\textbf{Generalization of ALIGN-generated interfaces generated with Qwen2.5-7B-Instruct to Llama3.3-70B-Instruct.}
    For each agent we report its score without the interface (w/o~ALIGN) and with the interface (w/~ALIGN); the value in parentheses is the \emph{absolute} improvement.}
    \label{tab:full_results_generalization_model_llama3_70b}
\end{table*}

\subsection{Generality Study Experiments\label{app:further}}

\begin{table*}[htbp]
    \centering
    \resizebox{\textwidth}{!}{%
    \begin{adjustbox}{max width=\linewidth}
    \begin{tabular}{@{}llrrrrrrr@{}}
        \toprule
        \textbf{Base Model} & \textbf{Interface} &
        \multicolumn{1}{c}{\textbf{\makecell{pick\\and place}}} &
        \multicolumn{1}{c}{\textbf{\makecell{pick clean\\and place}}} &
        \multicolumn{1}{c}{\textbf{\makecell{pick heat\\and place}}} &
        \multicolumn{1}{c}{\textbf{\makecell{pick cool\\and place}}} &
        \multicolumn{1}{c}{\textbf{\makecell{look at /\\examine\\in light}}} &
        \multicolumn{1}{c}{\textbf{\makecell{pick two obj\\and place}}} &
        \multicolumn{1}{c}{\textbf{\makecell{Success\\Rate (\%)}}} \\
        \midrule
        \multirow{2}{*}{gpt-4.1-mini}
        & w/o ALIGN & 58.33 & 22.58 & \leavevmode\hphantom{0}8.70 & \leavevmode\hphantom{0}9.52 & 22.22 & 52.94 & 28.36 \\
        & w/ ALIGN  & 95.83 & 87.10
                    & 26.09 & 80.95 
                    & 27.78  & 52.94 
                    & \textbf{64.93}  \\
        \midrule
        \multirow{2}{*}{gpt-4.1}
        & w/o ALIGN & 100.00 & 93.55 & 13.04 & 71.43 & 61.11 & 100.00 & 73.88 \\
        & w/ ALIGN  & 100.00 & 100.00 
                    & 78.26 & 100.00 
                    & 77.78  & 100.00 
                    & \textbf{93.28}  \\
        \bottomrule
    \end{tabular}
    \end{adjustbox}%
    }
    \caption{Experimental results for GPT-4.1 series agents with ALIGN on ALFWorld.}
    \label{tab:generality_3}
\end{table*}

For the validation on closed-source LLMs, we selected the GPT-4.1 family, experimenting with gpt-4.1-mini-2025-04-14 and gpt-4.1-2025-04-14.
Different from the main experiments, this setting also uses a cheaper generation stack: gpt-4.1-mini-2025-04-14 replaces GPT-4.1 both as the Analyzer and as the experimental-verification model of the Optimizer, while Gemini\,2.5\,Pro is still used for interface code synthesis.
We used a gpt-4.1-mini-based Vanilla agent to synthesize the interface with ALIGN, and then applied the same interface to agents powered by gpt-4.1-mini and gpt-4.1.
All other experimental settings were identical to those in the main experiments.
The results are presented in Table~\ref{tab:generality_3}; the same generation stack is analyzed further in Appendix~\ref{app:generator_spectrum}.

For domain-specific models trained within the environment, we used GiGPO-Qwen2.5-7B-Instruct-ALFWorld, a state-of-the-art model post-trained on ALFWorld via reinforcement learning~\citep{DBLP:journals/corr/abs-2505-10978}.
We reused the interface produced in our main experiment (generated with the base Qwen2.5-7B-Instruct model under the Vanilla Agent method).
At evaluation time, we considered two configurations: (1) our Vanilla Agent setting, and (2) a configuration that matches the logic and prompt setting used during training in the original paper.

\subsection{Ablation Study Experiments\label{app:ablation_results}}

The full result of interface ablation experiment can be found in Table~\ref{tab:full_results_ablation}.

\begin{table*}[htbp]
\small
\centering
    \begin{adjustbox}{max width=\linewidth}
    \begin{tabular}{@{}llcccc@{}}
    \toprule
    \textbf{Method} & \textbf{Interface} & \multicolumn{2}{c}{\textbf{Embodied}} & \multicolumn{1}{c}{\textbf{Web}} & \multicolumn{1}{c}{\textbf{Tool}} \\
    \cmidrule(lr){3-4} \cmidrule(lr){5-5} \cmidrule(lr){6-6}
    & & \textbf{ALFWorld} & \textbf{ScienceWorld} & \textbf{Webshop} & \textbf{M$^3$ToolEval} \\
    \midrule
    
    \multirow{3}{*}{Vanilla} & Full & 60.45 & 27.69 & 61.23 & 20.83 \\
    & w/o \textsc{InferRules} & 51.49 \scriptsize\textcolor{Green4}{(-8.96)} & 24.34 \scriptsize\textcolor{Green4}{(-3.35)} & 51.03 \scriptsize\textcolor{Green4}{(-10.20)} & 18.06 \scriptsize\textcolor{Green4}{(-2.77)} \\
    & w/o \textsc{WrapStep} & 26.87 \scriptsize\textcolor{Green4}{(-33.58)} & 22.97 \scriptsize\textcolor{Green4}{(-4.72)} & 61.23 \scriptsize{(-0.00)} & 11.11 \scriptsize\textcolor{Green4}{(-9.72)} \\
    \midrule
    
    \multirow{3}{*}{ReAct} & Full & 63.43 & 28.97 & 42.93 & 18.06 \\
    & w/o \textsc{InferRules} & 58.21 \scriptsize\textcolor{Green4}{(-5.22)} & 26.89 \scriptsize\textcolor{Green4}{(-2.08)} & 35.97 \scriptsize\textcolor{Green4}{(-6.96)} & 9.72 \scriptsize\textcolor{Green4}{(-8.34)} \\
    & w/o \textsc{WrapStep} & 45.52 \scriptsize\textcolor{Green4}{(-17.91)} & 22.53 \scriptsize\textcolor{Green4}{(-6.44)} & 47.60 \scriptsize\textcolor{red}{(+4.67)} & 19.44 \scriptsize\textcolor{red}{(+1.38)} \\
    \midrule
    
    \multirow{3}{*}{Self-Consistency} & Full & 69.40 & 25.41 & 61.10 & 16.67 \\
    & w/o \textsc{InferRules} & 67.91 \scriptsize\textcolor{Green4}{(-1.49)} & 23.11 \scriptsize\textcolor{Green4}{(-2.30)} & 55.67 \scriptsize\textcolor{Green4}{(-5.43)} & 13.89 \scriptsize\textcolor{Green4}{(-2.78)} \\
    & w/o \textsc{WrapStep} & 23.13 \scriptsize\textcolor{Green4}{(-46.27)} & 14.82 \scriptsize\textcolor{Green4}{(-10.59)} & 60.67 \scriptsize\textcolor{Green4}{(-0.43)} & 15.28 \scriptsize\textcolor{Green4}{(-1.39)} \\
    \midrule
    
    \multirow{3}{*}{Self-Refine} & Full & 40.30 & 22.99 & 52.30 & 6.94 \\
    & w/o \textsc{InferRules} & 32.84 \scriptsize\textcolor{Green4}{(-7.46)} & 21.27 \scriptsize\textcolor{Green4}{(-1.72)} & 46.33 \scriptsize\textcolor{Green4}{(-5.97)} & 6.94 \scriptsize{(-0.00)} \\
    & w/o \textsc{WrapStep} & 5.97 \scriptsize\textcolor{Green4}{(-34.33)} & 15.40 \scriptsize\textcolor{Green4}{(-7.59)} & 47.80 \scriptsize\textcolor{Green4}{(-4.50)} & 6.94 \scriptsize{(-0.00)} \\
    \midrule
    
    \multirow{3}{*}{Planning} & Full & 52.99 & 26.34 & 54.67 & 18.06 \\
    & w/o \textsc{InferRules} & 42.54 \scriptsize\textcolor{Green4}{(-10.45)} & 25.56 \scriptsize\textcolor{Green4}{(-0.78)} & 48.18 \scriptsize\textcolor{Green4}{(-6.49)} & 16.67 \scriptsize\textcolor{Green4}{(-1.39)} \\
    & w/o \textsc{WrapStep} & 26.12 \scriptsize\textcolor{Green4}{(-26.87)} & 16.48 \scriptsize\textcolor{Green4}{(-9.86)} & 52.87 \scriptsize\textcolor{Green4}{(-1.80)} & 16.67 \scriptsize\textcolor{Green4}{(-1.39)} \\
    \bottomrule
    \end{tabular}
    \end{adjustbox}%
\caption{Ablation study on the components of ALIGN. Values represent task success rates (\%) or scores. For ablated conditions (w/o \textsc{InferRules}, w/o \textsc{WrapStep}), performance changes from the `Full' are shown in parentheses.}
\label{tab:full_results_ablation}
\end{table*}

\section{Additional Analyses}
\label{app:additional}

\subsection{Positioning against Rule-Learning and Prompt-Optimization Methods\label{app:positioning}}

\begin{table*}[htbp]
\small
\centering
\renewcommand{\arraystretch}{1.15}
\begin{adjustbox}{max width=\linewidth}
\begin{tabular}{@{}p{2.6cm}p{3.8cm}p{3.3cm}p{4.8cm}@{}}
\toprule
 & \textbf{WALL-E / AutoManual} & \textbf{APO methods} & \textbf{ALIGN} \\
\midrule
What is optimized & Agent-side rules or manual & The agent's prompt & The environment's information and feedback functions ($f_{\text{info}}$, $f_{\text{obs}}$) \\
Artifact & Natural-language knowledge injected into the acting agent & Prompt text & Executable interface code (\textsc{InferRules} and \textsc{WrapStep}) \\
Verification & Rules induced from collected trajectories & Held-out scoring of candidate prompts & Controlled experiments executed in the environment before any change is accepted \\
Runtime feedback seen by the agent & Raw observations, unchanged & Raw observations, unchanged & Rewritten observations (e.g., ``Nothing happens'' $\rightarrow$ the violated precondition) \\
\bottomrule
\end{tabular}
\end{adjustbox}
\caption{Positioning of ALIGN against agent-side rule learning (WALL-E, AutoManual) and automatic prompt optimization (APO) methods.}
\label{tab:positioning}
\end{table*}

Table~\ref{tab:positioning} summarizes the boundary between ALIGN and the two families of methods that are structurally closest to it.
WALL-E~\citep{DBLP:journals/corr/abs-2410-07484} aligns the agent's internal world model by learning rules from collected trajectories, and AutoManual~\citep{DBLP:conf/nips/0001LYYL024} builds a manual that conditions its planner; APO methods~\citep{zhou2023ape,pryzant2023protegi,yang2024opro,wang2024promptagent} search over the agent's prompt with an LLM-driven optimizer and select candidates by held-out scoring.
All of them leave the environment's feedback channel, the source of the state aliasing studied in this paper, untouched, whereas ALIGN repairs the channel itself.
Two consequences follow.
First, the artifact is reusable below the agent abstraction: one interface serves five agent architectures (Table~\ref{tab:generalization}(a)), four model backbones (Table~\ref{tab:generalization}(b)), and an RL-specialized policy at evaluation time (Table~\ref{tab:generality}(b)).
AutoManual's manual also transfers between models, but as prompt content for a compatible planner, whereas the ALIGN wrapper needs no agent-side compatibility at all.
Second, the Analyzer--Optimizer loop is not APO with different variables: its search space is environment-side code, and its acceptance signal is executed experiments rather than a held-out score (Appendix~\ref{app:apis}).
Table~\ref{tab:sim-ablation} shows that this signal is load-bearing, as success collapses across iterations without it.

On the absolute numbers reported by these works (95\% on ALFWorld for WALL-E and 97.4\% for AutoManual): they rest on GPT-3.5/GPT-4-class base models together with agent-side scaffolds, whereas our main table uses Qwen2.5-7B-Instruct.
Under a comparable base model, a vanilla agent with the ALIGN-generated interface reaches 93.28\% (Table~\ref{tab:interface_comparison_1}), above the best agentic system we tested (AgentSquare, 86.57\%).
Because the two directions are complementary rather than competing, we directly test their combination in Appendix~\ref{app:complementarity}.

\subsection{Complementarity with Agent-Side Rule Learning\label{app:complementarity}}

\textbf{Setup.}~
We use the ten rules that AutoManual's GPT-4-based Builder evolved in its original ALFWorld experiments, as shipped in its public repository, and inject them into the agent prompt exactly as AutoManual's testing stage consumes a manual.
Owing to differences in ALFWorld versions and implementations, four mechanical adaptations were applied with the semantics left untouched: (1)~the Python helper \texttt{find\_object} was replaced by an equivalent textual description of the search procedure; (2)~``put X in/on Y'' was rewritten as ``move X to Y'' to match our action space; (3)~references to AutoManual's build log (e.g., ``as observed in epoch\_14'') were removed; and (4)~the naming style was changed from \texttt{desk\_1} to ``desk 1''.
The base models are Qwen2.5-7B-Instruct and Qwen3-8B~\citep{qwen3}, both under the Vanilla agent; Qwen3-8B reuses the interface generated with the Qwen2.5-7B-Instruct agent, i.e., the cross-model transfer setting of Table~\ref{tab:generalization}(b).
All other settings follow the main experiments.

\textbf{Results.}~
As shown in Table~\ref{tab:complementarity}, the combination is the best configuration for both model families (64.93\% and 53.73\%), and the two methods repair different levels of failure.
AutoManual's rules operate at the plan level, encoding task-type recipes, whereas ALIGN operates at the execution level, telling the agent why a step failed and which precondition applies.
Qwen2.5-7B-Instruct mostly knows what to do but fails in execution, as its issued actions diverge from its intent, so ALIGN contributes more (+47.02 vs.\ +15.67 points); Qwen3-8B shows the reverse profile (+32.09 vs.\ +36.57 points).
We also observed cross-family behavioral drift that affects the transferred interface on Qwen3-8B: for instance, Qwen2.5 emits ``use desklamp 1'', which the interface recognizes, whereas Qwen3 emits ``use desklamp 1 on X'', which it does not.
The transfer nevertheless remains strong (+32.09 points on Qwen3-8B), and generating more robust, family-general interfaces is a direction for future work.

\subsection{Content Audit of Injected Observations\label{app:content_audit}}

\begin{table}[htbp]
\small
\centering
\setlength{\tabcolsep}{4pt}
\resizebox{\columnwidth}{!}{%
\begin{adjustbox}{max width=\linewidth}
\begin{tabular}{lrrrr}
\toprule
\textbf{Domain} & \textbf{Mechanic} & \textbf{Borderline} & \textbf{Task-specific} & \textbf{Total}\\
\midrule
ALFWorld       & 116 & 6 & 0  & 122\\
WebShop        & 20  & 0 & 0  & 20\\
ScienceWorld   & 71  & 6 & 2  & 79\\
M$^3$ToolEval  & 72  & 0 & 19$^\dagger$ & 91\\
\midrule
\rowcolor{blue!4}
\textbf{Total} & 279 & 12 & 21 & 312\\
\bottomrule
\end{tabular}
\end{adjustbox}%
}
\caption{Content audit of all agent-facing strings injected by the 20 final interfaces. $^\dagger$The 19 M$^3$ToolEval strings instantiate about three distinct patterns.}
\label{tab:content_audit}
\end{table}

Interfaces are generated exclusively from training tasks and held fixed during testing (Section~\ref{sec:experimental_settings}), and the w/o and w/ ALIGN conditions share an identical protocol, so no test-set information enters generation.
The stricter question, specific to our own claim that the interface should reveal capability rather than substitute for the reasoning a benchmark measures, is whether the injected content clarifies environment mechanics or over-guides the agent.
We audited this directly.

\textbf{Criterion.}~
An injection is \emph{mechanic-level} iff it is (i)~a deterministic function of the current state and the attempted action, i.e., information the environment computed but failed to express, and (ii)~free of task-goal content: it never points to the target, spells out a solution procedure, or evaluates answers.
Strings that contain only generic hints naming no task entity (e.g., exploration nudges) but are not strictly determined by the state and action are marked \emph{borderline}; strings that violate (ii) are marked \emph{task-specific guidance}.
We extracted all 312 agent-facing strings from the 20 final interfaces (four benchmarks $\times$ five agent methods) and labeled each one; the audit data and scripts are released with the code.

\textbf{Findings.}~
Table~\ref{tab:content_audit} shows the result.
(1)~ALFWorld and WebShop, which are behind the headline results, contain no task-specific guidance; the six borderline ALFWorld strings are generic exploration nudges that name no task entity.
(2)~The flagged minority in ScienceWorld and M$^3$ToolEval splits into two kinds.
One kind is correctness feedback (e.g., ``Incorrect, try another''): the generated code computes it at run time from task metadata exposed by the environment, verbalizing a signal that the environment already computes internally; we regard these as environment-design repairs rather than violations, consistent with our finding that richer feedback should be prioritized by environment designers.
The other kind encodes procedure-level knowledge for a task type (e.g., an ordering of decoding tools, or a recipe for mixing paint).
Such rules are learned from training tasks only and generalize as code, so agent-side learners such as manuals or RL policies would legitimately carry the same knowledge; nevertheless, under our stricter claim we count them as over-guidance.
(3)~The procedure-level cases correlate exactly with the smallest training splits (one or two tasks per task type, Appendix~\ref{app:environments}), where the distinction between mechanics and procedures blurs for the Optimizer, and they occur in the two lowest-gain domains.

\textbf{Aggregate evidence.}~
Two further checks indicate that the gains are not over-guidance in aggregate.
The Valid Actions baseline (Table~\ref{tab:interface_comparison_2}) gives strictly more overt help, namely the full set of legal actions at every turn, yet reaches 44.03\% against 60.45\% for ALIGN; and the trajectory analysis in Appendix~\ref{app:behavior} shows failure-to-success conversions with preserved action entropy and increased exploration.

\textbf{Mitigation.}~
We add a task-agnosticism constraint to the Optimizer prompt (requirement~9 in Appendix~\ref{app:prompt}), which rejects injections that reference task-goal entities or spell out full procedures, and we flag the affected interfaces in the released audit data.
The interfaces evaluated in this paper were generated before this constraint was introduced.

\subsection{Exploration and Strategy Diversity\label{app:behavior}}

A natural concern is that explicitly disclosing implicit rules may undermine the agent's autonomous exploration or collapse its strategy diversity.
We therefore analyze the complete evaluation trajectories on ALFWorld and ScienceWorld for all five agent methods; as a sanity check, the same parser reproduces the reductions in consecutive invalid actions reported in Table~\ref{tab:iface-error}.
The analysis supports four observations.

\textbf{(1)~Diversity is preserved.}~
Action-type entropy stays in the same band with and without ALIGN (ALFWorld Vanilla 2.62 vs.\ 2.72 bits; ScienceWorld Vanilla 3.22 vs.\ 3.42) and is sometimes higher with ALIGN (ALFWorld ReAct 2.53 vs.\ 2.30).
Unique action trigrams are comparable on ALFWorld and higher with ALIGN on ScienceWorld (Vanilla 14.1 vs.\ 7.6).
The w/o ALIGN runs exhibit more raw action types, but the surplus consists largely of malformed actions emitted while stuck in invalid loops, i.e., noise rather than strategy.

\textbf{(2)~Exploration increases.}~
If rule disclosure substituted for exploration, agents would visit less of the environment; the opposite holds.
The mean number of distinct ``go to'' targets per task rises for every method (ALFWorld Vanilla 5.4 vs.\ 3.4, Self-Consistency 8.6 vs.\ 3.1, Self-Refine 4.6 vs.\ 1.0; ScienceWorld Vanilla 2.6 vs.\ 1.3).
\textsc{WrapStep} tells the agent why an action failed, never what to do next.

\textbf{(3)~Gains are failure-to-success conversions, not smoothing.}~
On ALFWorld, 50--79 test tasks per method are converted from failure to success against only 1--7 regressions; on ScienceWorld the corresponding numbers are 22--46 and 0--9.
On tasks solved in both conditions, ALFWorld agents use fewer steps with ALIGN (17.7 vs.\ 34.0 for Vanilla) because the tail of invalid loops disappears, whereas on ScienceWorld they take more valid actions to reach higher partial-credit scores.

\textbf{(4)~Content.}~
Together with the content audit in Appendix~\ref{app:content_audit}, these results indicate that the improvements stem from restored observability rather than from lowered task difficulty.

\subsection{Generator Capability Spectrum\label{app:generator_spectrum}}

\begin{table*}[htbp]
\small
\centering
\renewcommand{\arraystretch}{1.15}
\begin{adjustbox}{max width=\linewidth}
\begin{tabular}{@{}p{2.2cm}p{2.2cm}p{2.2cm}p{7.9cm}@{}}
\toprule
\textbf{Analyzer} & \textbf{Verification} & \textbf{Code synthesis} & \textbf{Outcome (ALFWorld, Vanilla agent)} \\
\midrule
GPT-4.1 & GPT-4.1 & Gemini\,2.5\,Pro & Main configuration (Table~\ref{tab:main_results}): Qwen2.5-7B-Instruct 13.43 $\rightarrow$ 60.45. \\
GPT-4.1-mini & GPT-4.1-mini & Gemini\,2.5\,Pro & Works: Qwen2.5-7B-Instruct 13.43 $\rightarrow$ 64.93 (no worse than the frontier stack); GPT-4.1-mini 28.36 $\rightarrow$ 64.93 and GPT-4.1 73.88 $\rightarrow$ 93.28 (Table~\ref{tab:generality_3}). \\
Qwen2.5-7B-Instruct & Qwen2.5-7B-Instruct & Qwen2.5-7B-Instruct & Fails to complete generation: across three runs with 32k/65k/131k context, the Analyzer never completes the verify-then-hand-off protocol (290 hypotheses, 0 completed verifications), so the Optimizer is never invoked and no interface is produced. \\
\bottomrule
\end{tabular}
\end{adjustbox}
\caption{Generator capability spectrum: the models driving the Analyzer, the experimental verification, and the interface code synthesis of the Optimizer, together with the resulting outcome.}
\label{tab:generator_spectrum}
\end{table*}

The main configuration uses frontier models in every generator role.
To characterize how much capability the generator actually needs, we vary the models driving its three roles (Table~\ref{tab:generator_spectrum}).

\textbf{A cheap model suffices for analysis and verification.}~
The Analyzer receives past interaction trajectories in its prompt, so it needs neither a strong world model nor prior knowledge of the environment.
Replacing both the Analyzer and the experimental-verification model of the Optimizer with GPT-4.1-mini yields interfaces that are no worse than those of the frontier stack: the interface generated with the Qwen2.5-7B-Instruct agent lifts it from 13.43\% to 64.93\% (vs.\ 60.45\% with the frontier stack), and the interface generated with the GPT-4.1-mini agent lifts GPT-4.1-mini from 28.36\% to 64.93\% and GPT-4.1 from 73.88\% to 93.28\%.
The demand for a frontier model therefore concentrates in a single step, the code synthesis of the Optimizer.

\textbf{A fully self-sufficient 7B stack fails.}~
With Qwen2.5-7B-Instruct in every role, generation does not complete.
Across three runs with 32k, 65k, and 131k context windows, which rules out a context-length artifact, the 7B Analyzer never completes the verify-then-hand-off protocol: it proposes 290 misalignment hypotheses but completes zero verifications, so the Optimizer is never invoked and no interface is produced.
The bottleneck is capability (structured tool use and exact-signature code generation), not context length.

\textbf{Takeaways.}~
The generator roles have a real capability floor, which we report as is; the floor matters little in practice, since generation is a one-time cost of about USD~9.4 per environment (Appendix~\ref{app:cost}), whereas the agent has no floor: the same 7B model that cannot generate an interface is the one the interface lifts from 13.43\% to 60.45\%.

\subsection{Audit of the Action Transducer\label{app:fact_audit}}

\begin{table}[htbp]
\small
\centering
\begin{adjustbox}{max width=\linewidth}
\begin{tabular}{@{}p{6.2cm}r@{}}
\toprule
\textbf{How the agent's action reaches the environment} & \textbf{Count}\\
\midrule
Passed verbatim (execute, then rewrite the observation) & 65\\
Read-only probe or navigation sub-steps added around it & 17\\
Intent-preserving surface normalization (same verb and target) & 5\\
Semantic remap to a different action ($f_{\text{act}}\neq$ identity) & 0\\
\midrule
\rowcolor{blue!4}
\textbf{Total \texttt{env.step} call sites in 20 interfaces} & \textbf{87}\\
\bottomrule
\end{tabular}
\end{adjustbox}
\caption{AST audit of all \texttt{env.step} call sites in the 20 deployed interfaces.}
\label{tab:fact_audit}
\end{table}

Section~\ref{sec:3.1} treats the action transducer $f_{\text{act}}$ as the identity mapping.
To verify that this scope is empirically appropriate rather than a bypass of intent-to-command translation, we extracted all 87 \texttt{env.step} call sites from the 20 deployed interfaces using the Python abstract syntax tree and classified how the agent's action reaches the environment (Table~\ref{tab:fact_audit}).
Every interface implements $f_{\text{info}}$ (20/20) and $f_{\text{obs}}$ (20/20), and none performs a semantic remap of the agent's action, whereas the human-designed interface in Appendix~\ref{app:interface_comparison} does (``converting put to move'').
In these environments, $f_{\text{act}}=$ identity is thus the empirically correct scope.
Environments with a genuine intent-to-command gap, where high-level goals must be compiled into low-level primitives, are the natural extension, and the formulation in Section~\ref{sec:3.1} already reserves this slot.

\section{Token Consumption Analysis}
\label{app:token}

\begin{table*}[htbp]
\small
\centering
\begin{adjustbox}{max width=\linewidth}
\begin{tabular}{llrrrr}
\toprule
& & \textbf{ALFWorld} & \textbf{ScienceWorld} & \textbf{WebShop} & \textbf{M$^3$ToolEval} \\
\midrule
\multirow{3}{*}{\textbf{Analyzer}}
  & Input Token (M)  & 0.2770 & 0.4333 & 0.1783 & 0.1094 \\
  & Output Token (M) & 0.0040 & 0.0036 & 0.0048 & 0.0016 \\
  & Total Token (M)  & 0.2809 & 0.4370 & 0.1831 & 0.1109 \\
\midrule
\multirow{3}{*}{\textbf{Optimizer}}
  & Input Token (M)  & 0.2619 & 0.2288 & 0.0669 & 0.1100 \\
  & Output Token (M) & 0.0087 & 0.0172 & 0.0040 & 0.0118 \\
  & Total Token (M)  & 0.2706 & 0.2460 & 0.0709 & 0.1217 \\
\midrule
\textbf{Total}
  & Total Token (M)  & 0.5515 & 0.6830 & 0.2540 & 0.2326 \\
\bottomrule
\end{tabular}
\end{adjustbox}
\caption{The average token consumption per iteration in the main experiment described in Section~\ref{sec:experimental_settings}.}
\label{tab:token}
\end{table*}

The average token consumption per iteration in the main experiment described in Section~\ref{sec:experimental_settings} is shown in Table~\ref{tab:token}.

Due to the ``Experimental Verification'' setup, the Analyzer and Optimizer need to interact with the environment multiple times, and all previous interaction histories are included as new prompt inputs to the LLM in each round of interaction. Additionally, when the Optimizer identifies that the generated interface is imperfect, it needs to refine the previously generated interface and conduct experimental verification again, leading to increased token consumption.
However, as LLM capabilities continue to improve and hallucination issues decrease, this cost will gradually reduce. Furthermore, it is worth noting that:

\begin{itemize}[left=0.4cm, itemsep=2pt, parsep=0pt]
    \item The \textsc{InferRules} wrapper and \textsc{WrapStep} wrapper are implemented as python logic code, which does not involve calls to models or agents, therefore not incurring additional token consumption. On the contrary, as demonstrated in our experiments in Section~\ref{sec:interface_quality}, \textbf{using ALIGN-generated interfaces can help agents reduce repetitive meaningless actions, thereby reducing the number of LLM calls and decreasing token consumption} compared to not using ALIGN-generated interfaces.
    \item Except when the Optimizer generates interface codes requiring the cutting edge LLMs (such as Gemini\,2.5\,Pro), weaker and more cost-effective LLMs (such as GPT-4.1-mini) can be used at other times, which will significantly reduce the operational costs of ALIGN. This is validated empirically in Appendix~\ref{app:generator_spectrum}.
    \item ALIGN-generated interfaces can generalize to different agent architectures and base LLMs. This means that for each environment, using the ALIGN method to generate an interface only once can bring performance improvements to different agents, regardless of agent version updates. This also means that \textbf{the cost of interface generation is a one-time expense}, rather than requiring the generation of new interfaces for each task execution. Therefore, from an amortization perspective, the method's cost becomes increasingly economical as the environment is utilized more frequently, with the one-time interface design cost being distributed across multiple uses and becoming proportionally smaller with increased usage.
\end{itemize}

\subsection{Cost Accounting\label{app:cost}}

\begin{table*}[htbp]
\small
\centering
\renewcommand{\arraystretch}{1.1}
\begin{adjustbox}{max width=\linewidth}
\begin{tabular}{@{}lccp{4.9cm}@{}}
\toprule
\textbf{Configuration} & \textbf{SR (\%)} & \textbf{One-time cost} & \textbf{Marginal cost per task}\\
\midrule
Qwen2.5-7B-Instruct (local), w/o ALIGN & 13.43 & -- & Local GPU only\\
Qwen2.5-7B-Instruct (local), w/ ALIGN  & 60.45 & $\approx$USD 9.4 & Local GPU only; 58\% of the baseline's input tokens and 43\% of its LLM calls\\
GPT-4.1-mini as base, w/o ALIGN        & 28.36 & -- & Per-task API cost, recurring\\
GPT-4.1 as base, w/o ALIGN             & 73.88 & -- & $\approx$USD 0.18 per task, recurring\\
GPT-4.1 as base, w/ ALIGN              & 93.28 & $\approx$USD 9.4 (same interface) & $\approx$58\% of the row above\\
\bottomrule
\end{tabular}
\end{adjustbox}
\caption{Cost comparison between an open-source base model equipped with the ALIGN-generated interface and closed-source models used directly as the base model, on ALFWorld with the Vanilla agent. Costs are computed from the evaluation logs and Table~\ref{tab:token} at list prices.}
\label{tab:cost}
\end{table*}

We quantify the cost of ALIGN from the evaluation logs and Table~\ref{tab:token}.

\textbf{Generation cost.}~
The 0.55M tokens per iteration on ALFWorld correspond to about USD~1.2 at list prices, so the full $K{=}8$ iterations cost about USD~9.4, and all four benchmarks together about USD~30.
This is a one-time cost per environment: the generated wrapper is model-free Python that is reused by every subsequent agent method, base model, and run.

\textbf{Inference savings.}~
The interface also reduces the steady-state cost of interaction.
From the evaluation logs, agents with ALIGN use 58\% of the baseline's input tokens and 43\% of its LLM calls per task; the mean per-task output drops from 353 to 142 tokens and the number of calls from 71.2 to 30.6.
The interface thus shrinks the interaction footprint rather than inflating it.

\textbf{Amortization.}~
Over the reuse in this paper (about 3,350 task-runs across agent methods and transferred backbones), generation costs about USD~0.003 per task-run; with a paid-API agent, it is recovered from inference savings alone after about 120 task-runs.
For reference, a single 134-task evaluation sweep of GPT-4.1 on ALFWorld (about USD~25) already exceeds the entire generation cost.

\textbf{Fair comparison with closed-source base models.}~
Table~\ref{tab:cost} compares Qwen2.5-7B-Instruct equipped with the ALIGN-generated interface against closed-source models used directly as the base model.
(i)~Qwen2.5-7B-Instruct with ALIGN more than doubles the success rate of the comparably affordable GPT-4.1-mini used as base (60.45\% vs.\ 28.36\%), with no recurring API cost.
(ii)~The choice is not either/or: the same interface applied to the closed-source base yields 93.28\%.

\subsection{Placement of Strong-Model Capability\label{app:placement}}

ALIGN is itself a collaboration scheme between a strong closed-source model and an open-source small model, with a deliberately chosen structure: the strong model works offline, once per environment, and its output is compiled into a reusable model-free artifact instead of being consulted online.
Two natural alternative placements of the same capability can be contrasted with it.
(1)~\emph{Online per-task teacher guidance} bills the strong model on every task, so the cost grows linearly with usage and the assistance is not reusable, whereas the cost of ALIGN is constant per environment (about USD~0.003 per task-run amortized) and its artifact transfers across five agent architectures and four model families (Table~\ref{tab:generalization} and Tables~\ref{tab:full_results_generalization_method}--\ref{tab:full_results_generalization_model_llama3_70b}).
(2)~\emph{Strong-model-built agent-side scaffolds} are compared in Appendix~\ref{app:interface_comparison}: AgentSquare configured with GPT-4.1 plus planning, reasoning, and memory modules reaches at best 86.57\% on ALFWorld, against 93.28\% for a vanilla agent with the ALIGN-generated interface (Table~\ref{tab:interface_comparison_1}), and interfaces hand-crafted by Ph.D.\ students reach 47.01\% against 60.45\% for ALIGN under the same Qwen2.5-7B-Instruct base (Table~\ref{tab:interface_comparison_2}).
The question is therefore not whether to add a strong model but where to spend its capability; one-time, verified, environment-side interface synthesis is empirically the highest-leverage placement among those we tested, and a matched-budget comparison against online-teacher schemes is left for future work.

\flushcolsend
\onecolumn
\raggedcolsend
\section{Implementation Details\label{app:experiment_settings}}

\subsection{Benchmarks Task Splits\label{app:environments}}

The task splits of benchmarks we use are as follows:

(1)~ALFWorld~\citep{DBLP:conf/iclr/ShridharYCBTH21}: We adhere to the original dataset partitioning presented in the paper, wherein the tasks from the ``eval\_out\_of\_distribution'' category are used as the test set, and the ``train'' category is designated as the training set. In each iteration, we randomly select three tasks from the training set of each task type to serve as the training data for the agent's interaction.

(2)~ScienceWorld~\citep{DBLP:conf/emnlp/WangJCA22}:We follow the original partitioning of the train and test sets as described in the paper. For efficiency reasons, during testing, we select at most the first five tasks from the 30 available task types for evaluation. In each iteration, we randomly select one task from the training set of each task type to be used as the training data for the agent's interaction.

(3)~WebShop~\citep{DBLP:conf/nips/Yao0YN22}: In alignment with the setup of \citet{DBLP:conf/iclr/YaoZYDSN023}, we use tasks with IDs ranging from 0 to 49 (50 tasks in total) as the test set, and tasks with IDs from 50 to 199 (150 tasks in total) as the training set. In each iteration, we randomly select 20 tasks from the training set to serve as the training data for the agent's interaction.

(4)~M$^{3}$ToolEval~\citep{DBLP:conf/icml/WangCY0L0J24}: Since M$^{3}$ToolEval does not provide a distinct training set division, we select two tasks from each task type in the original dataset as the training set, with the remaining tasks used as the test set. In each iteration, the entire training set is utilized for the agent's interaction.

\subsection{Hyperparameter and Experimental Setting}

For all the agents, we deploy them uniformly using vllm~\citep{kwon2023efficient} across 8 Nvidia A100 80GB GPUs, with the inference temperature set to 0.0. The models utilized contain Qwen2.5-7B-Instruct\footnote{\url{https://huggingface.co/Qwen/Qwen2.5-7B-Instruct}}~\citep{qwen2.5}, Qwen2.5-14B-Instruct\footnote{\url{https://huggingface.co/Qwen/Qwen2.5-14B-Instruct}}~\citep{qwen2.5}, Llama3.1-8B-Instruct\footnote{\url{https://huggingface.co/meta-llama/Llama-3.1-8B-Instruct}}~\citep{llama3.1} and Llama3.3-70B-Instruct\footnote{\url{https://huggingface.co/meta-llama/Llama-3.3-70B-Instruct}}~\citep{llama3.3}.

In ALIGN, we use Gemini\,2.5\,Pro~(gemini-2.5-pro-exp-03-25)\citep{google2025gemini25pro} for Optimizer to generate new interface, with the temperature set to 0.2. For other scenarios requiring the use of an LLM, we employ GPT-4.1~(gpt-4.1-2025-04-14)\citep{gpt4.1}. We set $K=8$ during experiments.

\subsection{Tools for Experimental Verification\label{app:apis}}

In order to implement the experimental verification process mentioned in Section~\ref{sec:ALIGN_framework}, we have encapsulated the following tools for Analyzer and Optimizer to interact with the interface-wrapped environment:

(1) \texttt{init\_simulator(task\_id, interface)}: Initializes an experimental task, specifying the task ID and the interface code.

(2) \texttt{reset\_simulator()}: Resets the experimental task.

(3) \texttt{run\_task()}: Runs the current task until completion, returning the interaction trajectory.

(4) \texttt{exec\_agent\_action(agent\_action)}: Executes a specific action and returns the enhanced observation after the interface processing.

(5) \texttt{get\_agent\_action()}: Based on the current trajectory, returns the next action to be issued by the agent.

(6) \texttt{change\_obs(obs)}: Modifies the observation of the previous action execution.

\subsection{Prompt Templates\label{app:prompt}}

We present the prompt template of the Analyzer and Optimizer for ALFWorld. For the prompt templates of other benchmarks, please refer to the supplemental materials. For the WebShop and M$^3$ToolEval environments, no ``Gold Action and Observation Sequence'' is provided. Requirement~9 of the Optimizer prompt for interface generation (the task-agnosticism constraint) was added after the content audit in Appendix~\ref{app:content_audit}.

\begin{tcolorbox}[alignprompt,title={Analyzer Prompt Template of Misalignment Analysis},title after break={Analyzer Prompt Template of Misalignment Analysis}]
\textbf{User message:} \par
In modern benchmarks evaluating LLM Agent reasoning capabilities, human designers create an Environment with a set of rules defining how tasks are accomplished. These rules, referred to as the Environment’s World Model, specify the sequence of actions required to achieve specific outcomes. For example, the Environment’s World Model might dictate that certain actions (e.g., operating on a receptacle) can only be performed after prerequisite actions (e.g., moving to the receptacle).\par
\par
Meanwhile, the Agent operates based on its own World Model, which it constructs by interpreting the task and environment prompts. The Agent first determines its high-level reasoning intent—its understanding of what needs to be done—and then selects actions according to its internal World Model. However, because the Environment’s World Model is manually crafted and may not be fully conveyed through prompts, the Agent’s World Model might differ, leading to unexpected behavior. For instance, the Agent might choose an action that aligns with its intent but violates the Environment’s rules, or it might misinterpret feedback due to insufficient information from the Environment.\par
\par
We define a misalignment between the Environment’s World Model and the Agent’s World Model as a situation where:\par
- The Environment provides feedback that does not sufficiently clarify its World Model, leaving the Agent unable to adjust its understanding of the rules.\par
\par
Your task is to analyze the logs from a recent task to determine whether such a misalignment occurred, preventing a fair assessment of the Agent’s capabilities. And this misalignment has not been fixed by current `WrapStep` function. Your analysis will guide us in addressing this issue moving forward.\par
\par
-----------------------------------------------------------------------\par
\#\#\# Experimental Environment Evaluation Template\par
\par
```python\par
\{\{ experimental\_template \}\}\par
```\par
\par
In this template, the function `InferRules` is used to define the environment rules. The function `WrapStep` handles post-processing of the Agent’s actions (e.g., splitting them into multiple steps, performing pre-checks, returning more detailed feedback, etc.). This function should not interfere with the Agent’s own reasoning. There current implementation is as follows:\par
\par
```python\par
\{\{ Interface \}\}\par
```\par
\par
-----------------------------------------------------------------------\par
\#\#\# Environment Logs\par
\par
```txt\par
\{\{ logs \}\}\par
```\par
\par
Here, each `Observation` is the feedback returned to the Agent after it executes an action.\par
\par
-----------------------------------------------------------------------\par
\#\#\# Gold Action and Observation Sequence\par
\par
```txt\par
\{\{ gold\_action\_obs\_sequence \}\}\par
```\par
\par
-----------------------------------------------------------------------\par
\#\#\# Environment Logics and Misalignment Analyzed in the Previous Steps\par
\par
\{\{ environment\_logics \}\}\par
\par
-----------------------------------------------------------------------\par
\#\#\# Your Task\par
\par
Determine whether, during this task, there was a misalignment between the Environment’s World Model and the Agent’s World Model that hindered a fair assessment of the Agent’s capabilities. Choose exactly one of the following outputs:\par
\par
If there is NO misalignment (i.e., the Agent’s failures stem from its own errors or limitations, not a mismatch with the Environment’s World Model), output:\par
<analysis\_result> No Misalignment </analysis\_result>\par
\par
If there IS a misalignment (i.e., the Environment’s World Model conflicts with the Agent’s World Model), output:\par
<analysis\_result> Found Misalignment </analysis\_result>\par
<environment\_logic\_and\_misalignments> the new environment rules and misalignments identified by you, which have not been fixed by current `WrapStep` function.\par </environment\_logic\_and\_misalignments>\par
\par
The format of the environment logic and misalignment is as follows:\par
```txt\par
\#\#\# Analysis Result 1\par
Analysis Task ID: xxx\par
Agent Action Type: xxx \# The type of action the Agent attempted to perform, such as "examine", "move object to receptacle", etc.\par
Agent Action Case: xxx \# The specific action the Agent attempted to perform.\par
Agent High-Level Reasoning Intent: xxx \# The Agent's high-level reasoning intent, which may be a general description of the action it was trying to perform.\par
Environment World Model Rule: xxx \# The rule from the Environment's World Model that don't align the Agent's World Model.\par
Sufficient Environment Feedback: xxx \# to offer the Agent adequate information to bridge gaps in understanding the environment's world model. such as "The environment should provide 'xxx' feedback when the Agent attempts to operate on a receptacle without first going to it."\par
Type: "Bug of current WrapStep function" or "Need to add new logic in the WrapStep function"\par
\par
\#\#\# Analysis Result 2\par
...\par
```\par
\par
Note: You should not generate duplicate misalignment analysis results as the ones already provided in the `Environment Logics and Misalignment Analyzed in the Previous Steps` section.\par
\end{tcolorbox}

\begin{tcolorbox}[alignprompt,title={Analyzer Prompt Template of Experimental Verification},title after break={Analyzer Prompt Template of Experimental Verification}]
\textbf{User message:} \par
Now you should conduct simulation experiments in the simulator to verify that the environment rules you hypothesized and Misalignment you identified truly exists. You must perform sufficient experiments to confirm or refute your suspicion.\par
\par
Here are the operations you can use:\par
\par
1. init\_simulator(task\_id: str)\par
   - Initializes a new simulator for the specified `task\_id`.\par
   - `task\_id` must be in the format 'int-int' where the first int $\in$ [0, 5].\par
   - The different task types are mapped as follows:\par
     {\par
       0: 'pick\_and\_place',\par
       1: 'pick\_clean\_and\_place',\par
       2: 'pick\_heat\_and\_place',\par
       3: 'pick\_cool\_and\_place',\par
       4: 'look\_at\_or\_examine\_in\_light',\par
       5: 'pick\_two\_obj\_and\_place'\par
     }\par
   - All subsequent operations occur within this initialized simulator.\par
\par
2. reset\_simulator()\par
   - Resets the current simulator to its initial state.\par
\par
3. execute\_agent\_action(agent\_action: str)\par
   - Executes an agent action using the `WrapStep` function.\par
\par
4. change\_last\_action\_observation(obs: str)\par
   - Updates the last observation returned by the simulator to the specified `obs`.\par
   - This is useful for simulating the agent’s next action in a different environment feedback context.\par
\par
5. get\_next\_agent\_action()\par
   - Retrieves the next action that the real Agent would perform under the current simulation conditions.\par
   - Note: The Agent’s choice of the next action is based on the current environment state, including the outcomes of any previous `step()` or `get\_next\_agent\_action()` call, along with the latest observations.\par
\par
If you believe you have reached a conclusion from your experiments, provide it in this format:\par
\par
<thought> Your reasoning here </thought>\par
<environment\_logic\_and\_misalignments> the new environment rules and misalignments identified by you, which have not been fixed by current `WrapStep` function. </environment\_logic\_and\_misalignments>\par
\par
The format of the environment logic and misalignment is as follows:\par
```txt\par
\#\#\# Analysis Result 1\par
Analysis Task ID: xxx\par
Agent Action Type: xxx \# The type of action the Agent attempted to perform, such as "examine", "move object to receptacle", etc.\par
Agent Action Case: xxx \# The specific action the Agent attempted to perform.\par
Agent High-Level Reasoning Intent: xxx \# The Agent's high-level reasoning intent, which may be a general description of the action it was trying to perform.\par
Environment World Model Rule: xxx \# The rule from the Environment's World Model that don't align the Agent's World Model.\par
Sufficient Environment Feedback: xxx \# to offer the Agent adequate information to bridge gaps in understanding the environment's world model. such as "The environment should provide 'xxx' feedback when the Agent attempts to operate on a receptacle without first going to it."\par
Type: "Bug of current WrapStep function" or "Need to add new logic in the WrapStep function"\par
\par
\#\#\# Analysis Result 2\par
...\par
```\par
\par
If you need to carry out more operations in the simulator, respond in the following format, specifying exactly one operation per turn:\par
\par
<thought> Your reasoning here, you should consider all hypotheses if the simulation result is not as expected </thought>\par
<action> The single operation you wish to perform (e.g., init\_simulator(task\_id="x-y"), step(action="x"), execute\_agent\_action(agent\_action="x"), etc.) </action>\par
\par
Note:\par
You should verify the correctness of the following, step by step, through your experiments:\par
1. environment\_rules: Use `execute\_agent\_action` to confirm that the environment rules you hypothesized are indeed correct, and current `WrapStep` function is not sufficient.\par
2. agent\_intent\_description: Obtain the Agent’s intended behavior (e.g., via `get\_next\_agent\_action`) and simulate it by using `WrapStep` to confirm whether it aligns with your description.\par
3. identified\_misalignment: Through chaning the environment feedback, you can verify whether the misalignment you identified is indeed correct and the environment feedback you hypothesized is indeed sufficient. You can use `WrapStep` to simulate the agent’s action, then use `change\_last\_action\_observation` to change the environment feedback, and finally use `get\_next\_agent\_action` to check whether the agent can correctly identify the next action.\par
\end{tcolorbox}

\begin{tcolorbox}[alignprompt,title={Analyzer Prompt Template of Reranking Misalignments Analysis (Ablation Study)},title after break={Analyzer Prompt Template of Reranking Misalignments Analysis (Ablation Study)}]
\textbf{User message:} \par
In modern benchmarks evaluating LLM Agent reasoning capabilities, human designers create an Environment with a set of rules defining how tasks are accomplished. These rules, referred to as the Environment’s World Model, specify the sequence of actions required to achieve specific outcomes. For example, the Environment’s World Model might dictate that certain actions (e.g., operating on a receptacle) can only be performed after prerequisite actions (e.g., moving to the receptacle).\par
\par
Meanwhile, the Agent operates based on its own World Model, which it constructs by interpreting the task and environment prompts. The Agent first determines its high-level reasoning intent—its understanding of what needs to be done—and then selects actions according to its internal World Model. However, because the Environment’s World Model is manually crafted and may not be fully conveyed through prompts, the Agent’s World Model might differ, leading to unexpected behavior. For instance, the Agent might choose an action that aligns with its intent but violates the Environment’s rules, or it might misinterpret feedback due to insufficient information from the Environment.\par
\par
We define a misalignment between the Environment’s World Model and the Agent’s World Model as a situation where:\par
- The Environment provides feedback that does not sufficiently clarify its World Model, leaving the Agent unable to adjust its understanding of the rules.\par
\par
Now other human experts have analyzed the logs from a recent task and identified some potential misalignments. Your task is to review these misalignments and choose the most appropriate one.\par
\par
-----------------------------------------------------------------------\par
\#\#\# Experimental Environment Evaluation Template\par
\par
```python\par
\{\{ experimental\_template \}\}\par
```\par
\par
In this template, the function `InferRules` is used to define the environment rules. The function `WrapStep` handles post-processing of the Agent’s actions (e.g., splitting them into multiple steps, performing pre-checks, returning more detailed feedback, etc.). This function should not interfere with the Agent’s own reasoning. There current implementation is as follows:\par
\par
```python\par
\{\{ Interface \}\}\par
```\par
\par
-----------------------------------------------------------------------\par
\#\#\# Environment Logs\par
\par
```txt\par
\{\{ logs \}\}\par
```\par
\par
Here, each `Observation` is the feedback returned to the Agent after it executes an action.\par
\par
-----------------------------------------------------------------------\par
\#\#\# Gold Action and Observation Sequence\par
\par
```txt\par
\{\{ gold\_action\_obs\_sequence \}\}\par
```\par
\par
-----------------------------------------------------------------------\par
\#\#\# Environment Logics and Misalignment Analyzed in the Previous Steps\par
\par
\{\{ environment\_logics \}\}
\
Note: These logics may not be accurate. They are the environment rules that were previously hypothesized and may contain errors.\par
\par
-----------------------------------------------------------------------\par
\#\#\# Your Task\par
\par
Choose the most appropriate misalignment analyzed by human experts from the list below:\par
\par
\{\{ new\_environment\_logics \}\}\par
\par
You should respond in format as follows:\par
```\par
<review> Your review of each expert output one by one </review>\par
<expert\_id> id of the selected expert output, only the number </expert\_id>\par
```\par
\end{tcolorbox}

\begin{tcolorbox}[alignprompt,title={Optimizer Prompt Template of Interface Generation},title after break={Optimizer Prompt Template of Interface Generation}]
\textbf{User message:} \par
In modern benchmarks evaluating LLM Agent reasoning capabilities, human designers create an Environment with a set of rules defining how tasks are accomplished. These rules, referred to as the Environment’s World Model, specify the sequence of actions required to achieve specific outcomes. For example, the Environment’s World Model might dictate that certain actions (e.g., operating on a receptacle) can only be performed after prerequisite actions (e.g., moving to the receptacle).\par
\par
Meanwhile, the Agent operates based on its own World Model, which it constructs by interpreting the task and environment prompts. The Agent first determines its high-level reasoning intent—its understanding of what needs to be done—and then selects actions according to its internal World Model. However, because the Environment’s World Model is manually crafted and may not be fully conveyed through prompts, the Agent’s World Model might differ, leading to unexpected behavior. For instance, the Agent might choose an action that aligns with its intent but violates the Environment’s rules, or it might misinterpret feedback due to insufficient information from the Environment.\par
\par
We define a misalignment between the Environment’s World Model and the Agent’s World Model as a situation where:\par
- The Environment provides feedback that does not sufficiently clarify its World Model, leaving the Agent unable to adjust its understanding of the rules.\par
\par
Your task is to refine the environment’s behavior based on the misalignment identified by the AnalysisAgent, ensuring the Agent’s true intentions are executed and its reasoning capabilities are fairly assessed.\par
\par
-----------------------------------------------------------------------\par
\#\#\# Experimental Environment Evaluation Template\par
\par
```python\par
\{\{ experimental\_template \}\}\par
```\par
\par
In this template, the function `InferRules` is used to define the environment rules. The function `WrapStep` handles post-processing of the Agent’s actions (e.g., splitting them into multiple steps, performing pre-checks, returning more detailed feedback, etc.). This function should not interfere with the Agent’s own reasoning. There current implementation is as follows:\par
\par
```python\par
\{\{ WrapStep \}\}\par
```\par
\par
-----------------------------------------------------------------------\par
\#\#\# Environment Logics and Misalignment Analyzed by AnalysisAgent Previously\par
\par
\{\{ last\_environment\_logics \}\}\par
\par
-----------------------------------------------------------------------\par
\#\#\# New Environment Logics and Misalignment Analyzed by AnalysisAgent\par
\par
\{\{ new\_environment\_logics \}\}\par
\par
-----------------------------------------------------------------------\par
\#\#\# Your Task\par
\par
Based on the misalignments identified by the AnalysisAgent, you need to refine and enhance the `InferRules` function and `WrapStep` function to align the Environment’s World Model with the Agent’s actions and provide clearer feedback. Your output should present the new versions of these functions, ensuring the Agent’s high-level reasoning intent is preserved.\par
Please ensure you follow these requirements:\par
\par
1. **Function Signature**  \par
   The function signature must be:\par
   ```python\par
   def InferRules(init\_obs, task)\par
     - init\_obs: str, the initial observation from the environment, containing all receptacles.\par
     - task: str, the task description.\par
\par
   def WrapStep(env, init\_obs, task, agent\_action: str, logger)\par
   ```\par
\par
2. **Return Values**\par
   The `InferRules` function’s return value must be a string that describes the environment rules.\par
\par
   The `WrapStep` function’s return value must be three items:\par
   ```python\par
   obs: str, reward: bool, done: bool\par
   ```\par
\par
3. **`env.step` Usage**  \par
   The only permitted usage pattern for `env.step` is:\par
   ```python\par
   obs, reward, done, info = env.step([agent\_action])\par
   obs, reward, done = obs[0], info['won'][0], done[0]\par
   ```\par
   No alternative usage forms are allowed. Each call to env.step causes an irreversible change to the environment state; actions must therefore be chosen carefully.\par
\par
4. **Package Imports**  \par
   You may import other packages if necessary, but you must include all imports in your code.\par
\par
5. **Multiple Calls and Conditional Returns**  \par
   You are free to call `env.step` multiple times or return different `obs` depending on `agent\_action` or the outcomes of these calls.\par
\par
6. **You can use logger.debug**\par
   You can use `logger.debug` to log any information you find useful. The logging will be captured and returned to you in the future for further analysis.\par
\par
7. Do not modify any aspects not explicitly identified by the AnalysisAgent in the “New Environment Logics and Misalignment Analyzed by AnalysisAgent” section.\par
\par
8. You must use the following approach when addressing the identified misalignment:\par
	- For each action defined in environment, provide clear, informative, and sufficient feedback from the environment whenever an invalid action is attempted, guiding the Agent toward understanding and adhering to the environment’s rules.\par
\par
9. **Task-Agnosticism Constraint**\par
   Every string returned by `InferRules` or injected into `obs` by `WrapStep` must describe environment mechanics only. It must be a deterministic function of the current environment state and the attempted action, and it must not reference task-goal entities, spell out a solution procedure for a task type, or evaluate whether the Agent’s answer is correct. Injections that violate this constraint will be rejected during experimental verification.\par
\par
10. **Output Format**  \par
   You must provide the output strictly in the following format:\par
   <thought>YOUR\_THOUGHT\_PROCESS\_HERE</thought>\par
   <code>YOUR\_CODE\_HERE</code>\par
\par
Please ensure your final answer follows these guidelines so that we can accurately bridge the misalignment and allow the environment to execute the Agent’s true intentions.\par
\end{tcolorbox}

\begin{tcolorbox}[alignprompt,title={Optimizer Prompt Template of Experimental Verification},title after break={Optimizer Prompt Template of Experimental Verification}]
\textbf{User message:} \par
Now you should conduct simulation experiments in the simulator to verify if the `InferRules` and `WrapStep` function you provided is correct for the new environment logics and misalignment analyzed by the AnalysisAgent.\par
\par
You must perform sufficient experiments to confirm or refute your suspicion. Here are the operations you can use:\par
\par
1. init\_simulator(task\_id: str)\par
   - Initializes a new simulator for the specified `task\_id`.\par
   - `task\_id` must be in the format 'int-int' where the first int $\in$ [0, 5].\par
   - The different task types are mapped as follows:\par
     {\par
       0: 'pick\_and\_place',\par
       1: 'pick\_clean\_and\_place',\par
       2: 'pick\_heat\_and\_place',\par
       3: 'pick\_cool\_and\_place',\par
       4: 'look\_at\_or\_examine\_in\_light',\par
       5: 'pick\_two\_obj\_and\_place'\par
     }\par
   - All subsequent operations occur within this initialized simulator.\par
\par
2. reset\_simulator()\par
   - Resets the current simulator to its initial state.\par
\par
3. execute\_agent\_action(agent\_action: str)\par
   - Executes an agent action using the `WrapStep` function you generated.\par
\par
4. change\_last\_action\_observation(obs: str)\par
   - Updates the last observation returned by the simulator to the specified `obs`.\par
   - This is useful for simulating the agent’s next action in a different environment feedback context.\par
\par
5. get\_next\_agent\_action()\par
   - Retrieves the next action that the real Agent would perform under the current simulation conditions.\par
   - Note: The Agent’s choice of the next action is based on the current environment state, including the outcomes of any previous `step()` or `get\_next\_agent\_action()` call, along with the latest observations.\par
\par
6. run\_task(task\_id: str)\par
   - Runs the entire task in the simulator and returns the running log.\par
   - After running the whole task, you need to call `init\_simulator` or `reset\_simulator` to reinitialize the simulator for further operations.\par
\par
If you believe you have reached a conclusion from your experiments, provide it in this format:\par
\par
<thought> Your reasoning here </thought>\par
<if\_need\_refine> True/False </if\_need\_refine>\par
<refine\_strategy> Your strategy for refining the WrapStep function, if if\_need\_refine is True </refine\_strategy>\par
\par
If you need to carry out more operations in the simulator, respond in the following format, specifying exactly one operation per turn:\par
\par
<thought> Your reasoning here, you should consider all hypotheses if the simulation result is not as expected </thought>\par
<action> The single operation you wish to perform (e.g., init\_simulator(task\_id="x-y"), step(action="x"), execute\_agent\_action(agent\_action="x"), etc.) </action>\par
\end{tcolorbox}

\begin{tcolorbox}[alignprompt,title={Optimizer Prompt Template of Reranking Interface Generation (Ablation Study)},title after break={Optimizer Prompt Template of Reranking Interface Generation (Ablation Study)}]
\textbf{User message:} \par
In modern benchmarks evaluating LLM Agent reasoning capabilities, human designers create an Environment with a set of rules defining how tasks are accomplished. These rules, referred to as the Environment’s World Model, specify the sequence of actions required to achieve specific outcomes. For example, the Environment’s World Model might dictate that certain actions (e.g., operating on a receptacle) can only be performed after prerequisite actions (e.g., moving to the receptacle).\par
\par
Meanwhile, the Agent operates based on its own World Model, which it constructs by interpreting the task and environment prompts. The Agent first determines its high-level reasoning intent—its understanding of what needs to be done—and then selects actions according to its internal World Model. However, because the Environment’s World Model is manually crafted and may not be fully conveyed through prompts, the Agent’s World Model might differ, leading to unexpected behavior. For instance, the Agent might choose an action that aligns with its intent but violates the Environment’s rules, or it might misinterpret feedback due to insufficient information from the Environment.\par
\par
We define a misalignment between the Environment’s World Model and the Agent’s World Model as a situation where:\par
- The Environment provides feedback that does not sufficiently clarify its World Model, leaving the Agent unable to adjust its understanding of the rules.\par
\par
Now other human experts have generated a set of code patches to address the misalignment between the Environment’s World Model and the Agent’s World Model. Your task is to evaluate these patches and select the best one.\par
\par
-----------------------------------------------------------------------\par
\#\#\# Experimental Environment Evaluation Template\par
\par
```python\par
\{\{ experimental\_template \}\}\par
```\par
\par
In this template, the function `InferRules` is used to define the environment rules. The function `WrapStep` handles post-processing of the Agent’s actions (e.g., splitting them into multiple steps, performing pre-checks, returning more detailed feedback, etc.). This function should not interfere with the Agent’s own reasoning. There current implementation is as follows:\par
\par
```python\par
\{\{ WrapStep \}\}\par
```\par
\par
-----------------------------------------------------------------------\par
\#\#\# Environment Logics and Misalignment Analyzed by AnalysisAgent Previously\par
\par
\{\{ last\_environment\_logics \}\}\par
\par
-----------------------------------------------------------------------\par
\#\#\# New Environment Logics and Misalignment Analyzed by AnalysisAgent\par
\par
\{\{ new\_environment\_logics \}\}\par
\par
-----------------------------------------------------------------------\par
\#\#\# Your Task\par
\par
Choose the best code from the following options to address the misalignment between the Environment’s World Model and the Agent’s World Model:\par
\par
\{\{ code\_patches \}\}\par
\par
You should respond in format as follows:\par
```\par
<review> Your review of each code one by one </review>\par
<code\_id> id of the selected result, only the number </code\_id>\par
```\par
\end{tcolorbox}

\clearpage
We present the prompt template of the Vanilla agent in ALFWorld to illustrate the usage of the $\textsc{InferRules}$. For the prompt templates of other agent methods and benchmarks, please refer to the supplemental materials.

\begin{tcolorbox}[alignprompt,title={Vanilla Agent Prompt Template in ALFWorld},title after break={Vanilla Agent Prompt Template in ALFWorld}]
\textbf{System message:} \par
You are an AI assistant solving tasks in a household environment. Your goal is to break down complex tasks into simple steps and plan your actions accordingly.\par
\par
\# Action Space\par
\par
In this environment, you have a set of high-level actions at your disposal, each corresponding to a typical household activity. These actions are:\par
\par
- look:                             look around your current location\par
- inventory:                        check your current inventory\par
- go to (receptacle):               move to a receptacle\par
- open (receptacle):                open a receptacle\par
- close (receptacle):               close a receptacle\par
- take (object) from (receptacle):  take an object from a receptacle\par
- move (object) to (receptacle):    place an object in or on a receptacle\par
- examine (something):              examine a receptacle or an object\par
- use (object):                     use an object\par
- heat (object) with (receptacle):  heat an object using a receptacle\par
- clean (object) with (receptacle): clean an object using a receptacle\par
- cool (object) with (receptacle):  cool an object using a receptacle\par
- slice (object) with (object):     slice an object using a sharp object\par
\par
Although each action may internally consist of multiple embodied steps (e.g., walking to the sink, turning a knob, etc.), from your perspective you need only provide one high-level action at a time.\par
\par
\# Instructions\par
\par
Single Action per Turn\par
At each step, you must respond with exactly one action (i.e., the next “thought”). Use the format:\par
ACTION [object/receptacle specifier]\par
ACTION [object/receptacle specifier]\par
For example:\par
take apple from table\par
or\par
go to kitchen.\par
\par
Environment Feedback\par
After you provide your single action, the environment will automatically execute it and return the resulting observation. You then decide on your next action based on the updated state.\par
\par
Reasoning (Chain of Thought)\par
You may use hidden reasoning to figure out the best next step. However, only output the single action that represents your decision. Do not reveal your entire chain of thought.\par
\par
Continue Until Task Completion\par
You will iterate this process—receiving the environment’s feedback, deciding on the next action, and outputting a single action—until the task is finished.\par
\par
\# Environment Rule\par
\par
\{InferRules(init\_obs, task)\}\par
\par
\textbf{User message:} \par
\# Task\par
\par
\{initial\_obs\}\par
\par
Begin by examining the environment or taking any initial steps you find relevant. Remember, provide only one action each time.\par
\end{tcolorbox}

\end{document}